\documentclass{article}
\usepackage[preprint]{neurips_2025}

\usepackage{url}
\usepackage{hyperref}
\usepackage{microtype}
\usepackage{graphicx}
\usepackage{booktabs} 
\usepackage{comment}

\usepackage{amsmath}
\usepackage{amssymb}
\usepackage{mathtools}
\usepackage{amsthm}

\usepackage{textcomp}
\usepackage{tikz}
\usepackage{pgfplots}
\usepackage{multirow}
\usepackage{xcolor}
\usepackage{listings}
\usepackage{listingsutf8}
\usepackage{mathpartir} 

\AtBeginDocument{%
  \newgeometry{
    textheight=9in,
    textwidth=6.5in,
    top=1in,
    headheight=12pt,
    headsep=25pt,
    footskip=30pt
  }%
}

\newif\ifdraftcomments
\draftcommentstrue 

\usetikzlibrary{shapes.geometric, arrows, positioning, fit, backgrounds, calc}
\pgfplotsset{compat=1.17}

\definecolor{inputcolor}{RGB}{52, 152, 219}    
\definecolor{processcolor}{RGB}{46, 204, 113}  
\definecolor{outputcolor}{RGB}{231, 76, 60}    
\definecolor{highlightcolor}{RGB}{155, 89, 182} 
\definecolor{phase1color}{RGB}{52, 152, 219}   
\definecolor{phase2color}{RGB}{46, 204, 113}   
\definecolor{phase3color}{RGB}{241, 196, 15}   
\definecolor{dslcolor}{RGB}{39, 60, 117}       
\definecolor{rulecolor}{RGB}{100, 100, 100}    

\lstdefinestyle{ucutlass}{
  basicstyle=\ttfamily\small,
  keywordstyle=\color{dslcolor}\bfseries,
  commentstyle=\color{rulecolor}\itshape,
  stringstyle=\color{processcolor},
  breaklines=true,
  columns=flexible,
  keepspaces=true,
  showstringspaces=false,
  literate={>>}{{$\gg$}}2,
  morekeywords={gemm,conv2d_fprop,conv3d_fprop,conv1d_fprop,batched_gemm,
    grouped_gemm,depthwise_conv2d,pipeline,transpose},
}

\begin{document}

\title{Improving Efficiency of GPU Kernel Optimization Agents using a Domain-Specific Language and Speed-of-Light Guidance}

\author{Siva Kumar Sastry Hari, Vignesh Balaji, Sana Damani, Qijing Huang, Christos Kozyrakis\\[4pt]
NVIDIA}

\maketitle

\begin{abstract}

Optimizing GPU kernels with LLM agents is an iterative process over a large design space. Every candidate must be generated, compiled, validated, and profiled, so fewer trials will save both runtime and cost. We make two key observations. First, the abstraction level that agents operate at is important. If it is too low, the LLM wastes reasoning on low-impact details. If it is too high, it may miss important optimization choices. Second, agents cannot easily tell when they reach the point of diminishing returns, wasting resources as they continue searching. 

These observations motivate two design principles to improve efficiency: (1) a compact domain-specific language (DSL) that can be learned in context and lets the model reason at a higher level while preserving important optimization levers, and (2) Speed-of-Light (SOL) guidance that uses first-principles performance bounds to steer and budget search. 
We implement these principles in $\mu$CUTLASS, a DSL with a compiler for CUTLASS-backed GPU kernels that covers kernel configuration, epilogue fusion, and multi-stage pipelines. We use SOL guidance to estimate headroom and guide optimization trials, deprioritize problems that are near SOL, and flag kernels that game the benchmark.

On 59 KernelBench problems with the same iteration budgets, switching from generating low-level code to DSL code using GPT-5-mini turns a $0.40\times$ geomean regression into a $1.27\times$ speedup over PyTorch. Adding SOL-guided steering raises this to $1.56\times$. Across model tiers, $\mu$CUTLASS + SOL-guidance lets weaker models outperform stronger baseline agents at lower token cost.
SOL-guided budgeting saves 19--43\% of tokens while retaining at least 95\% of geomean speedup, with the best policy reaching a $1.68\times$ efficiency gain. Lastly, SOL analysis helps detect benchmark-gaming cases, where kernels may appear fast while failing to perform the intended computation.

\end{abstract}

\section{Introduction}
\label{sec:introduction}

Training and inference in modern machine learning systems rely heavily on GPUs, so GPU kernel performance directly affects end-to-end runtime.
PyTorch provides a strong baseline, but writing kernels in CUDA or open libraries such as CUTLASS~\cite{cutlass2024} that match or beat it is still hard and requires specialized expertise.
The challenge is compounded by the pace of GPU evolution. New generations introduce performance-critical features roughly every year or two. Without AI assistance many developers may not finish optimizing for one GPU generation before the next becomes available.
So, Large Language Model (LLM) agents are not merely a helpful automation tool, but a necessity as applications and hardware feature sets evolve faster. 

GPU kernel optimization is an iterative loop. It involves proposing a candidate, compiling it against a host-side driver, testing for correctness, and profiling.
Time to solution depends on the number of iterations the agent takes to converge to a correct, fast kernel and can increase further if the system-side steps consume more time.
%
%
For one vanilla-agent run on our system on a single problem (with 40 generate-compile-test-profile attempts), roughly 21\% of the total time is spent in LLM calls versus 79\% in compile+run+profile tool actions. While the exact breakdown will vary based on factors such as the target system, problem size, and LLM inference latency, reducing the number of iterations will help reduce the latency to solution. Moreover, spending more tokens through additional iterations does not necessarily yield a proportionately faster solution.
KernelBench~\cite{ouyang2025kernelbench} reports that LLM-generated CUDA agents often fail to improve over compiled PyTorch baselines, and that corraborates with our experiments where the GPT-5-mini based agent that tries to generates CUDA/CUTLASS code regresses relative to PyTorch, despite passing the compilation step at high rates. Even when LLM-based optimization approaches do succeed~\cite{ouyang2025kernelbench, sakana2024cuda, romera2024funsearch}, they often require many iterations to do so.
Therefore, the objective we choose is iteration efficiency, i.e., the speedup we gain per generate--compile--test--profile cycle.

One source of inefficiency in CUDA agents is the abstraction level they operate at. When generating raw CUDA or CUTLASS code (or code for any other open-source library),\footnote{In this paper we focus on open-source kernel generation, following the broader KernelBench goal of synthesizing open alternatives to library kernels rather than invoking additional closed-source implementations.} the LLM must simultaneously choose an effective optimization strategy and correctly emit many lines of complex (or template-heavy) C++, navigating architecture-specific alignment constraints, scheduler compatibility rules, and layout algebra. This burden grows as GPU architectures evolve rapidly across generations, introducing new features (e.g., TMA, NVFP4) that are increasingly important to exploit for high-performance kernels, while models may lag behind the latest toolchains and programming abstractions. From a programming languages perspective, the output representation strongly influences how often an expensive compile-test-and-profile pass yields informative feedback. A representation that factors out implementation boilerplate and statically rules out invalid configurations early, rather than at runtime, lets the LLM spend less effort on code-generation correctness and more on performance-critical choices. This motivates a design question: what is the ideal abstraction level for an LLM-driven GPU kernel optimization agent?

Another source of inefficiency is the lack of directional guidance and a principled estimate of remaining headroom beyond the current implementation. While agents use profiling, profiling provides only a local view of the current implementation. It cannot tell the agent whether substantially better performance may still be achievable through different implementation or algorithmic choices. 
As a result, the agent does not know whether it should keep optimizing the current implementation or switch to a different strategy, nor does it know whether spending more resources on this problem is likely to be worthwhile. This lack of global headroom information hurts both within-problem steering and across-problem budget allocation.

These two challenges motivate two complementary design principles to improve the efficiency of LLM-driven kernel optimization agents.
First, we introduce a high-level domain-specific language (DSL) to address the abstraction problem by raising the representation from raw code to a compact, statically validated specification that exposes the main optimization levers an LLM must reason about. A key requirement is that the DSL remain small enough to be learned entirely in context while still being expressive enough to capture high-impact choices in the target domain and optimization space.
We instantiate this principle as $\mu$CUTLASS, a DSL for CUTLASS-backed GPU kernels. $\mu$CUTLASS lets the agent specify high-level kernel choices while a compiler generates the corresponding low-level CUTLASS implementation.
The DSL covers four facets of GPU kernel optimization: operator selection, kernel configuration (tile shapes, scheduling policies, data types, cluster dimensions), epilogue fusion (composing activations, bias, scaling, and custom expressions into the kernel epilogue via the \texttt{>>} operator), and multi-stage pipelines (orchestrating layout transforms with fused dtype conversion alongside kernel stages).
Its grammar is AI-generated with a human in the loop and remains compact enough to be learned in context. Our EBNF specification is about 170 lines, including constraint annotations enforced by the compiler.

Second, we use first-principles performance analysis (SOL analysis) to quantify room for performance improvement and use this to steer optimization search, allocate resources across problems, and to discover reward hacking (or ``gaming").
We integrate SOL guidance in three ways. For a given problem, we estimate theoretical SOL performance from the problem's memory and compute requirements together with the target hardware specifications. This provides a quantitative headroom signal that helps the agent reason about and prioritize optimization hypotheses.
We call this SOL-first workflow \emph{MANTIS}. It \textit{Measures} the current performance, performs SOL \textit{Analysis}, \textit{Nominates} and \textit{Triages} optimization hypotheses before \textit{Implementing} them, and finally \textit{Summarizes} the results and consolidates learnings in long-term memory.
Across problems, SOL enables smarter budget scheduling. Once a problem is sufficiently close to its SOL bound or stagnates, additional attempts have diminishing returns, so the scheduler can reallocate effort to problems with more remaining headroom. Finally, SOL supports integrity checking in two ways. We use it both as a strict runtime bounds check and as additional specification for an LLM-based reviewer about what work the kernel is expected to perform, which helps detect reward hacking (or gaming) attempts with implementations that skip the intended computation but pass the correctness harness.

We evaluated $\mu$CUTLASS and SOL guidance on 59 KernelBench problems\footnote{KernelBench has 250 GPU kernel problems across Levels~1--3. We use a 59-problem subset selected to include problems that dominate modern and emerging text LLM workloads (Transformers, SSMs, recurrent models) based on profiling representative models, and exclude pooling and legacy CV/RNN tasks.} on H100 across three models: GPT-5-mini, GPT-5, and GPT-5.2.
Whereas a naive GPT-5-mini agent shows a $0.40\times$ regression over PyTorch, the $\mu$CUTLASS agent achieves a $1.27\times$ geomean speedup with the same iteration budget. The DSL helps across all three model tiers, with the largest gains on the weakest model.
SOL-guided steering further improves the performance of GPT-5-mini to $1.56\times$ and GPT-5 to $2.07\times$ over Pytorch. Crucially, with our solution, GPT-5-mini + $\mu$CUTLASS + SOL-guidance outperforms the GPT-5 baseline, and GPT-5 + $\mu$CUTLASS + SOL-guidance matches the GPT-5.2 baseline at a lower inference cost.

Across problems, SOL-guided budget scheduling saves 19--43\% of tokens while retaining at least 95\% of geomean speedup across all evaluated configurations. The best configuration reaches a $1.68\times$ efficiency gain with 43\% token savings and 96\% speedup retention. We found that integrity checking is essential because LLMs can exploit specification loopholes and produce kernels that run fast  without actually doing the intended work.  SOL analysis helps here by providing a runtime ceiling to check against and by augmenting the specification for the LLM-based reviewer with a structured description of the required work, potentially reducing loopholes. Without integrity filtering, reported speedups would inflate by up to $1.9\times$.

\section{Related Work}
\label{sec:related}

\subsection{Compiler Autotuning and Scheduling Systems}

ML compilers and autotuners treat kernel optimization as search over constrained program or schedule spaces, guided by measurements, cost models, and adaptive search policies.
Halide~\cite{ragan2013halide} introduced the separation of algorithm from schedule, enabling systematic search over optimization strategies.
OpenTuner~\cite{ansel2014opentuner} generalized the idea of composing multiple search strategies, while TVM~\cite{chen2018tvm} and Ansor~\cite{zheng2020ansor} advanced learned cost models and automated schedule search for tensor programs. MetaSchedule~\cite{shao2022metaschedule} continues this line with a modular probabilistic search space over tensor-program transformations.
These systems share a core insight with $\mu$CUTLASS, i.e., exposing optimization choices through a compact DSL makes search more productive than unconstrained code generation.
Our setting adds two requirements that autotuning does not face.
First, the search agent is an LLM with no prior exposure to the DSL, so the language must be compact enough to be learnable entirely in-context.
Second, each attempt triggers a compile/run/profile cycle that adds latency (especially for larger problems), so static validation should reject invalid configurations before the time-consuming toolchain runs.
An interesting future direction is to expose autotuners themselves as tools or skills to the LLM, combining semantic reasoning with more efficient search over local schedule spaces.

\subsection{GPU Programming DSLs and Kernel Libraries}

CUTLASS~\cite{cutlass2024}, built on CuTe tensor/layout templates~\cite{cute2024}, provides high-performance GEMM and convolution templates but exposes a large configuration surface---hundreds of template parameters spanning data types, layouts, tile shapes, scheduling policies, and pipeline stages---that can be difficult for an LLM to emit correctly and consistently.
Triton-style DSLs~\cite{tillet2019triton} reduce boilerplate and accelerate kernel development, but peak performance on the newest architectures is frequently realized via CUDA/CUTLASS kernels that encode architecture-aware kernel designs (e.g., persistent, asynchronous pipelines)~\cite{wright2024pingpong, pytorch2024tma}.
Similar trends appear in state-of-the-art Hopper attention and mixed-input GEMM kernels implemented with CUTLASS/CuTe~\cite{pytorch2024flashattn3, wilkinson2024machete}.
$\mu$CUTLASS, in contrast, is not a general kernel programming language.
Instead, it exposes only the performance-critical knobs of the CUTLASS template library while hiding template instantiation, header management, and architecture-specific plumbing.
This design trades expressiveness for reliability in iterative optimization settings minimizing wasted time-consuming tool actions.
The underlying principle---factoring a library's performance-critical choices into a compact, statically validated DSL that an LLM can 
learn in-context---is not specific to CUTLASS and generalizes to other target libraries/languages and kernel-authoring IR ecosystems (e.g., Triton and MLIR dialects)~\cite{tillet2019triton, lattner2020mlir}.

\subsection{Performance Bounds}

The roofline model~\cite{williams2009roofline} introduced bound-and-bottleneck analysis as a visual framework for reasoning about whether a kernel is compute-bound or memory-bound, given peak hardware throughput.
This idea has since been applied broadly, including to LLM inference workloads~\cite{yuan2024roofline}, and extended with simulation/mapping-exploration-based analyses that derive attainable data movement and compute intensity bounds with some buffer capacity constraints~\cite{huang2024orojenesis}.

\subsection{LLMs for Kernel Generation and Optimization}

With $\mu$CUTLASS, we do not change the search algorithm, we change the search space as we raise the level of abstraction. Our SOL-guidance loop (MANTIS) uses first-principles bounds to steer and budget exploration, and can be paired with different local search procedures (e.g., beam search with varying parallelism and rollout depth).

LLM-based code generation and tool-using agents are now common~\cite{li2022competition, li2023starcoder, roziere2023code, yang2024swe, wang2024openhands, shinn2023reflexion}. We focus on GPU kernel optimization, where each attempt incurs an expensive compile/run/profile loop and low-level code generation can waste iterations.

KernelBench~\cite{ouyang2025kernelbench} establishes an evaluation framework and shows that iterative refinement with execution feedback is necessary for competitive performance, but reports that raw LLM-generated CUDA frequently fails to improve over compiled PyTorch.Autocomp~\cite{hong2025autocomp} combines planning with beam search for accelerator kernels.
Sakana AI's Evolving CUDA Engineer~\cite{sakana2024cuda} applies evolutionary search to CUDA optimization.
FunSearch~\cite{romera2024funsearch}, AlphaTensor~\cite{fawzi2022alphatensor}, and AlphaDev~\cite{mankowitz2023alphadev} explore evolutionary or reinforcement-learning-based program discovery.
Deep RL has also been applied directly to CUDA kernel generation: CUDA-L1~\cite{li2025cuda} uses contrastive reinforcement learning to improve kernel optimization, CUDA Agent~\cite{dai2026cuda} scales agentic RL for high-performance kernel generation, and K-Search~\cite{cao2026k} maintains a co-evolved intrinsic world model to guide search.
Recent concurrent work instead targets progress per tool action.
GPU Kernel Scientist~\cite{andrews2025gpukernelscientist} uses profiling signals to generate targeted hypotheses.
Astra~\cite{wei2025astra} studies multi-agent controllers.
Kevin~\cite{baronio2025kevin} explores multi-turn reinforcement learning, and minimal executable programs~\cite{chu2025mep} reduce feedback cost.
GEPA and TextGrad~\cite{agrawal2025gepa, yuksekgonul2024textgrad} optimize agent instructions from textual execution feedback.
Most of these methods search more effectively within existing code-level representations or controller loops.
In contrast, we change the representation via $\mu$CUTLASS and use SOL bounds to reason about remaining headroom, stopping, and budget reallocation, which are complementary to advances in search algorithms and controller design.

As agents take longer optimization traces, context management becomes a first-order design variable.
Recent work studies how to construct and curate the agent context (what to keep, what to summarize, and what to retrieve), including Agentic Context Engineering~\cite{zhang2025ace}, retrieval-augmented generation for self-improvement~\cite{wang2024dynamic}, and explicit memory mechanisms~\cite{packer2023memgpt, madaan2022memprompt}.
In parallel, In-Context Reinforcement Learning (ICRL) formalizes adaptation at inference time from interaction histories (without weight updates)~\cite{moeini2025survey, monea2024llmicrl, wang2024tdtransformers, krishnamurthy2024explore, dong2026kernelblaster}.
In this study, we adopt a lightweight form of context management by passing forward only limited information from prior attempts; this choice is largely orthogonal to $\mu$CUTLASS and SOL-guidance.

\section{$\mu$CUTLASS: A DSL for CUTLASS GPU Kernels}
\label{sec:ucutlass}

$\mu$CUTLASS is a compact DSL that lets an agent specify a set of CUTLASS kernel choices without directly generating CUTLASS code.
More broadly, we design $\mu$CUTLASS around the following three design goals:

\begin{itemize}

\item \emph{Compact and learnable in-context.}
Assume the model has no prior exposure to the DSL. The core grammar and a few examples must fit in a short prompt and be learnable in-context. The model should be able to use them to write high-quality programs.

\item \emph{Support abstract reasoning and high-level code generation.}
The DSL should let the model reason about higher-level kernel optimization choices and make larger design changes without getting distracted by low-level scaffolding, template details, or other minor implementation details.
This shift away from low-level details toward higher-level choices can lead to larger performance gains.

\item \emph{Retain high-impact control choices.}
The target language (e.g., CUTLASS) exposes a large space of valid programs and parameters.
We want a small set of high-impact control choices that still preserves the main knobs needed to get close to SOL, but without exposing the full target language.
High-impact control choices are named decisions where a small change in the source program can introduce a significant change in performance (e.g., data type, layout, schedule, tiling).
The DSL should also provide simple ways to express the common patterns the agent must explore, such as fused epilogues or an explicit pipeline boundary.

\end{itemize}

Following these design goals, we design $\mu$CUTLASS as a compact DSL to author CUTLASS kernels.
The DSL factors out boilerplate and prevents common structural errors (e.g., template mismatches, architecture gating, alignment rules).
Agents (or humans) write short specifications. The compiler validates them and generates a CUTLASS C++ implementation.
$\mu$CUTLASS is not a general GPU programming language. It is a compact interface to CUTLASS that exposes the decisions that matter while hiding template instantiation and scaffolding.
It covers CUTLASS's main operator families and extends the vocabulary with operations and compositions that occur frequently in LLM workloads.
The goal is not full coverage, but to capture the parts of the design space that drive most performance variation while keeping the language small enough to learn in-context.
It supports three core features:

\begin{itemize}
\item \emph{Kernel configuration}: operation, data types, layouts, tile shapes, scheduling policy, alignment, and pipeline stages.
\item \emph{Epilogue fusion}: compose common epilogue operations with \texttt{>>}, compiled to a single fused epilogue.
\item \emph{Multi-stage pipelines}: compose non-fusable stages with explicit kernel boundaries using \texttt{pipeline(...)}.
\end{itemize}

\begin{figure*}[t]
  \centering
  \resizebox{\textwidth}{!}{%
    \begin{tikzpicture}[
  font=\normalsize,
  node distance=6mm and 9mm,
  box/.style={draw, rounded corners, inner sep=5pt, text width=#1, align=left},
  box/.default=34mm,
  inbox/.style={box=#1, fill=inputcolor!8, minimum height=44mm},
  inbox/.default=70mm,
  step/.style={box=#1, fill=processcolor!8, align=left, minimum height=24mm},
  step/.default=34mm,
  outbox/.style={box=#1, fill=outputcolor!7, align=left, minimum height=44mm},
  outbox/.default=92mm,
  detour/.style={box=#1, fill=highlightcolor!7, align=left},
  detour/.default=60mm,
  arrow/.style={-latex, line width=0.8pt},
  darrow/.style={-latex, dashed, line width=0.8pt},
]

  \node[inbox] (dsl) {%
    \textbf{Input: SM90a GEMM + Epilogue Fusion ({\ttfamily kernel.dsl})}\\[2pt]
    {\ttfamily\footnotesize
    gemm()\\
    \ \ .with\_dtype(input=fp16, acc=fp32, output=fp16)\\
    \ \ .with\_layout(A=RowMajor, B=RowMajor, C=RowMajor)\\
    \ \ .with\_arch(sm\_90a)\\
    \ \ .with\_threadblockshape(m=128, n=128, k=64)\\
    \ \ .with\_stages(2)\\
    \ \ .with\_alignment(A=8, B=8, C=8)\\
    \ \ .with\_scheduler(kernel=tma, epilogue=tma)\\
    \ \ >> bias() >> gelu() >> clip(min=0, max=6)
    }\\[2pt]
    \footnotesize (Also supports {\ttfamily pipeline(\dots)} for explicit kernel boundaries.)%
  };

  \node[step=46mm, right=of dsl] (grammar) {%
    \textbf{Parse and Lower}\\[-1pt]
    \footnotesize Lark grammar ({\ttfamily grammar.lark})\\
    \footnotesize syntax check + parse tree\\
    \footnotesize $\to$ configuration IR\\
    {\ttfamily\footnotesize (KernelConfig / PipelineConfig)}%
  };

  \node[step=46mm, right=of grammar] (check) {%
    \textbf{Static Validation}\\[-1pt]
    \scriptsize Architecture constraints: dtype/layout/features\\
    \scriptsize schedule/alignment/stages checks\\
    \scriptsize op/fusion checks %
  };

  \node[outbox, right=of check] (emit) {%
    \textbf{Emit Code + Wrapper}\\[-1pt]
    \footnotesize deterministic namespace + PyTorch wrapper entrypoints\\
    \footnotesize sha256(config) $\to$ {\ttfamily ucutlass\_\textit{hash}}\\[2pt]
    {\ttfamily\footnotesize ucutlass\_\textit{hash}.h}\\[-1pt]
    {\ttfamily\footnotesize
    namespace ucutlass\_\textit{hash} \{\\
    \ \ // ... CUTLASS includes + type aliases ...\\
    \ \ /* Original DSL: gemm().with\_dtype(...).with\_arch(sm\_90a) ... */\\
    \ \ using DeviceKernel = cutlass::gemm::device::GemmUniversalAdapter<\dots>;\\
    \ \ at::Tensor kernel\_kernel(A, B, /*C*/, alpha, beta);\\
    \ \ // ... pack strides + build arguments + allocate workspace\\
    \ \ // ... run on current PyTorch CUDA stream; return output tensor\\
    \ \ at::Tensor kernel\_kernel\_ex(/*extended params*/);\\
    \}\  // namespace ...\\
    using ucutlass\_\textit{hash}::kernel\_kernel;\\
    }%
  };

  \node[detour, below=20mm of grammar, xshift=-20mm] (pycutlass) {%
    \textbf{Emit CUTLASS Python API ({\ttfamily kernel.py})}\\[1pt]
    {\ttfamily\footnotesize
    plan = cutlass.op.Gemm(\dots)\\
    plan.epilogue = evt.chain(bias(), gelu(), clip(\dots))\\
    cpp = plan.emit()
    }%
  };

  \node[detour, right=of pycutlass] (cppgen) {%
    \textbf{{\ttfamily cutlass\_cppgen}}\\[-1pt]
    \footnotesize compile plan $\to$ C++ kernel code%
  };

  \node[detour, right=of cppgen] (cppout) {%
    \textbf{Generated {\ttfamily kernel.cu}}\\[1pt]
    {\ttfamily\footnotesize
    using Gemm = cutlass::gemm::device::GemmUniversal<\dots>;\\
    // ... epilogue visitor tree ...\\
    }%
  };

  \draw[arrow] (dsl) -- (grammar);
  \draw[arrow] (grammar) -- (check);
  \draw[arrow] (check) -- (emit);

  \draw[darrow] (check.south) -- node[font=\scriptsize, midway, sloped, above,
    fill=white, inner sep=1pt] {optional} (pycutlass.north);
  \draw[arrow] (pycutlass) -- node[font=\scriptsize, above] {send} (cppgen);
  \draw[arrow] (cppgen) -- node[font=\scriptsize, above] {emit} (cppout);
  \draw[darrow] (cppout.north) -- node[font=\scriptsize, midway, right] {edit + wrap} (emit.south);

\end{tikzpicture}}
  \caption{$\mu$CUTLASS compilation pipeline. The compiler parses and lowers a \texttt{kernel.dsl} program to a typed configuration IR, validates architecture and other constraints, and emits the CUTLASS C++ code in a wrapper header (\texttt{ucutlass\_hash.h}). Optionally, it emits a CUTLASS Python API plan (\texttt{kernel.py}) and invokes \texttt{cutlass\_cppgen} to produce \texttt{kernel.cu}, which is then wrapped into the final header.}
  \label{fig:ucutlass-pipeline}
\end{figure*}
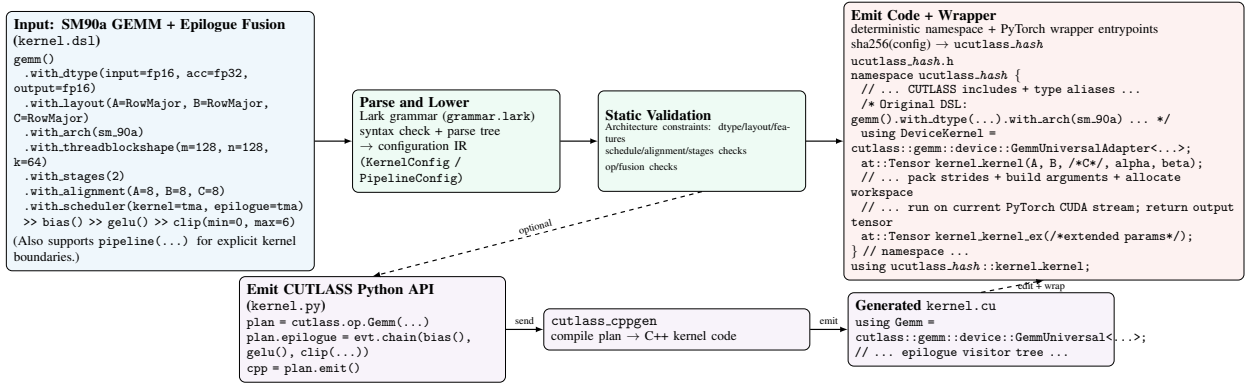

All choices are explicit and named, which avoids hidden defaults that an agent (or human) must guess.
Most kernels are short (often $\sim$10--20 lines), simplifying the agent's job by shifting boilerplate and validity checks to the compiler.
This raises the abstraction for \emph{agent-driven} custom kernel generation: once the DSL specification is accepted, much of the kernel-level validation work is eliminated, and what remains is to verify that the agent integrates the generated kernel correctly into the full program.
Figure~\ref{fig:ucutlass-pipeline} includes a concrete \texttt{kernel.dsl} example (a GEMM with fused epilogue operations) alongside the compilation pipeline.
Table~\ref{tab:ucutlass-coverage} summarizes the supported operator families and feature support.
It includes GEMM (standard, batched, grouped), 2D/3D convolutions (forward, data gradient, weight gradient), 1D convolutions (via 2D lowering with $H=1$), depthwise convolutions (via a CuTe backend on SM90+), and grouped convolutions (SM80--89). 
Data types include FP64, FP32, FP16, BF16, FP8 (SM90+), and INT8.

\begin{table*}[t]
\centering
\scriptsize
\setlength{\tabcolsep}{2.5pt}
\renewcommand{\arraystretch}{1.08}
\begin{minipage}[t]{0.53\textwidth}\vspace{0pt}
\centering
\textbf{(a) Operator coverage}\\[-1pt]
\begin{tabular}{l l p{25mm}}
\hline
Operation family & Arch support & Notes / restrictions \\
\hline
GEMM & SM70+ & --- \\
Grouped GEMM & SM80+ & --- \\
Conv2d & SM70+ & NHWC \\
Conv3d & SM70+ & NDHWC \\
Conv3d wgrad & SM70--89 & SM90+ $\times$ \\
Conv1d & SM70+ & Lowered to Conv2d, $H{=}1$ \\
Depthwise Conv & SM70--89; SM90+$^*$ & CuTe backend on SM90+ \\
Grouped Conv & SM80--89 & --- \\
\hline
\end{tabular}

\vspace{3mm}
\textbf{(c) Supported epilogues} (\texttt{>>}) \\
\begin{tabular}{p{0.98\linewidth}}
\hline
Built-in: relu, gelu, silu, sigmoid, tanh, mish, hardswish, leaky\_relu, elu, clip, clamp, bias, per\_channel\_scale, per\_row\_scale, per\_col\_scale, scale, aux\_store, aux\_load \\
Custom: \texttt{custom('expr', inputs=\{\dots\})} (SM90a$^\dagger$) \\
\hline
\end{tabular}
\end{minipage}\hfill%
\begin{minipage}[t]{0.46\textwidth}\vspace{0pt}
\centering
\textbf{(b) Feature support}\\[-1pt]
\begin{tabular}{l l}
\hline
Feature / binding & Arch support \\
\hline
\texttt{.with\_dtype(\dots)} & SM70+ \\
\texttt{.with\_arch(\dots)} & SM70+ \\
\texttt{.with\_alignment(\dots)} & SM70+ \\
\texttt{.with\_stages(\dots)} & SM70+ \\
\texttt{.with\_tile} & SM70--89 \\
\texttt{.with\_threadblockshape} & SM90+ \\
\texttt{.with\_cluster} & SM90+ \\
\texttt{.with\_scheduler} & SM90+ \\
\texttt{.with\_swizzle} & SM70--89 \\
\texttt{.with\_iterator} (conv) & SM70--89 \\
\texttt{.with\_split\_k} (conv) & SM70--89 \\
\texttt{pipeline}/\texttt{transpose} & SM70+ \\
Transpose dtype conversion & SM70+ \\
\texttt{custom(\ldots)} epilogues & SM90a$^\dagger$ \\
\texttt{.with\_operand\_swap(true)} & SM90+$^\ddagger$ \\
\hline
\end{tabular}
\end{minipage}
\caption{Coverage and feature support in $\mu$CUTLASS. $^*$On SM90+, depthwise convolutions route to a CuTe-based backend with additional restrictions (e.g., limited stride/dilation and epilogue support). $^\dagger$Custom epilogue expressions are supported on SM90a. $^\ddagger$\texttt{.with\_operand\_swap(true)} applies to SM90+ FP32 GEMM and requires $M{=}N$.}
\label{tab:ucutlass-coverage}
\end{table*}

\paragraph{Why a new DSL?}
Existing GPU kernel authoring systems expose large and expressive design spaces, but that expressiveness is often a poor fit for LLM-driven iterative optimization. Among them, CUTLASS/CuTe are publicly available libraries that provide an expansive feature set to write near-SOL kernels for many common deep learning operators on latest GPU architectures. However, they also expose a rich set of knobs and require substantial low-level, template-heavy code~\cite{cutlass2024, cute2024}. Triton reduces boilerplate, but still requires synthesizing full kernels and reasoning about low-level program structure and memory access~\cite{tillet2019triton}.
$\mu$CUTLASS is one instantiation following our design goals. Any representation that meets these goals could serve the same role, including as a compact version of an existing kernel language.

\paragraph{Compilation.}
Once the user or the agent writes the $\mu$CUTLASS program, the compiler parses it, lowers it to a typed configuration IR, and checks the constraints (e.g., architecture-level compatibility and alignment conditions) before emitting any C++.
When validation fails, we try to explain what went wrong and why, so the model can often fix the specification before triggering a compile/run/profile attempt.
For well-formed programs, the compiler routes to an appropriate code-generation backend and emits a CUTLASS template instantiation plus a PyTorch-compatible driver. On SM90+ GEMMs, we emit directly through the CUTLASS~3.x CollectiveBuilder API; on SM70--89 operations and convolutions, we emit via the CUTLASS Python API (\texttt{cutlass\_cppgen}) and wrap the generated code into the final header.
Each generated file is placed in a deterministic namespace derived from a hash of the configuration (as shown in Figure~\ref{fig:ucutlass-pipeline}), and the original $\mu$CUTLASS source is embedded as a comment for traceability, enabling caching and reliable comparisons across attempts.
\texttt{pipeline(\ldots)} programs compile to multi-stage drivers that run explicit transform stages (layout transposes fused with dtype conversion), then the main kernel stage, and then optional transforms to switch back to the original layout and dtype.
This is useful when a kernel expects a different layout or dtype than the input and output tensors in the model.
On SM90+, we fuse epilogue chains into a single Epilogue Visitor Tree (EVT)~\cite{chen2024evt}.
The default \texttt{cutlass\_cppgen} flow does not directly emit C++ for EVT, so we extend it to enable EVT in $\mu$CUTLASS.

\paragraph{Implementation and testing.}
We developed the grammar and compiler with AI plus human-in-the-loop iteration, translating the design principles above into a standalone, testable tool.
The compiler is implemented as a Python CLI that supports both file-based compilation and a tool-oriented mode that accepts a DSL program as text, which we use to expose to the agent runtime.
We validate the compiler and DSL independently of KernelBench by building a standalone example suite. Each example in this suite provides the DSL code and a driver C++ code to instantiate the kernel and checks correctness against PyTorch references.

A natural next research direction is to auto-generate a compact DSL and validate it from a target set of operations and target backend code, so we can retarget this approach beyond CUTLASS.

\section{Speed-of-Light (SOL) Guidance}
\label{sec:agent_design}

\subsection{SOL Analysis}
\label{sec:sol_analysis}

For each problem, we perform a Speed-of-Light (SOL) analysis using a roofline-style model of the full reference computation, which may consist of a single operator or a multi-operator graph.
The analysis proceeds in the following four steps.

\begin{itemize}

\item \emph{Problem characterization} identifies the operators, their dimensions, and data types, and estimates total FLOPs and best-case DRAM bytes, assuming each unique input element is read once and each output is written once, with fusion of intermediates where feasible.

\item \emph{Hardware limits} are obtained from the GPU's peak compute throughput and memory bandwidth from published specifications, scaled by the current clock frequencies reported by \texttt{nvidia-smi} or a default 1500 MHz.

\item \emph{Roofline bound} computes the arithmetic intensity (FLOPs/byte), the compute-bound time $T_\mathrm{compute} = \text{FLOPs} / \text{Peak~FLOP/s}$, the memory-bound time $T_\mathrm{mem} = \text{Bytes} / \text{Peak~BW}$, and the lower-bound runtime $t_\mathrm{SOL} = \max(T_\mathrm{compute}, T_\mathrm{mem})$.

\item \emph{Bottleneck classification} compares the arithmetic intensity to the ridge point to determine whether the workload is compute-bound or memory-bound.

\end{itemize}

The result is a compact structured report that the agent can use for downstream steering, scheduling, and integrity checking.
In this study, the SOL report is generated by an LLM. An example report is included in Appendix~\ref{sec:appendix:sol-report}.
It can also be produced by an analytical system such as Orojenesis~\cite{huang2024orojenesis} or SOLAR~\cite{nvlabs2025solar}.

The tightness of the resulting SOL bound depends on the modeling assumptions.
Not accounting for how data maps to on-chip buffers and caches can make the SOL bound too tight. The current model assumes perfect caching, but if on-chip storage cannot hold all reused data, the kernel must re-fetch from slower memory, so the true runtime will be slower than the estimate.
Conversely, not accounting for reduced precision or input-value sparsity can make the bound too loose. A kernel that uses FP16 or FP8 arithmetic, for example, can perform the same work faster than an FP32-based estimate because the hardware provides higher throughput for reduced-precision arithmetic.
In this study, KernelBench problems are specified with FP32 tensors, but agents are allowed to use reduced-precision math. For optimization steering, we use the LLM-generated SOL estimate under an FP32 problem formulation with TF32 throughput assumptions. 
For budget scheduling and integrity checking, we use an FP16-based estimate, since optimized kernels may use FP16 arithmetic and this provides a tighter ceiling for their achievable runtime. 
Inputs and outputs remain FP32, as allocated.
KernelBench problems initialize tensors with random values, making sparsity exploitation unlikely, so we do not model it.

\subsection{SOL-Guided Optimization Hypotheses Exploration}
\label{sec:agent:within_problem}

Within a single problem, we use the SOL report together with the current best implementation's measured runtime to quantify the remaining SOL gap and steer hypothesis generation.
Let $t_\mathrm{best}$ denote the measured time of the current best correct kernel (fastest so far) and $t_\mathrm{SOL}$ denote the SOL bound.
The ratio $g = t_\mathrm{best} / t_\mathrm{SOL}$ gives a quantitative gap signal: the larger $g$ is, the further the current implementation is from SOL. We use this signal along with Nsight Compute profiling data to nominate and prioritize optimization hypotheses that target the dominant performance gaps.
We call this SOL-first workflow \emph{MANTIS}, which operates in six phases: \textit{M}easure--\textit{A}nalyze--\textit{N}ominate--\textit{T}riage--\textit{I}mplement--\textit{S}ummarize.\footnote{
MANTIS is inspired by the Assess--Choose--Execute (ACE) Helix and the Observe--Orient--Decide--Act (OODA) loop used in the fighter-jet community for decades\cite{lee2023clearthinking} .
In this analogy, \emph{Measure} and \emph{Analyze} correspond to \emph{Assess} in ACE and to \emph{Observe}+\emph{Orient} in OODA. \emph{Nominate} and \emph{Triage} correspond to \emph{Choose} in ACE and to \emph{Decide} in OODA. \emph{Implement} corresponds to \emph{Execute}/\emph{Act}.
Finally, we add \emph{Summarize} to capture post-iteration learning, analogous to post-mission debriefs.
}

\begin{enumerate}
\item \textit{Measure}: Profile the current best kernel via Nsight Compute (\texttt{ncu}) to obtain occupancy, memory and compute throughput, and other metrics.

\item \textit{Analyze}: Perform the SOL analysis, as described above in Section~\ref{sec:sol_analysis}, and combine it with the profiler summary and current source code to quantify the SOL gap ($g = t_\mathrm{best} / t_\mathrm{SOL}$) and attribute it to specific bottlenecks. In practice, the SOL report can be computed once during bootstrap and reused across iterations.

\item \textit{Nominate}: Generate multiple optimization hypotheses with causal reasoning linking each proposal to the analyzed bottleneck(s) and explaining how it is expected to close the SOL gap.

\item \textit{Triage}: Rank hypotheses by a \emph{gap-aware} return-on-investment (ROI) formula that amplifies ambition when far from SOL and encourages incremental gains when close to it:
\[
\mathrm{ROI}(h) = \frac{(\widehat{S}(h))^{1 + \max(0,\,\log_{10}(g/5))}}{\widehat{R}_\mathrm{impl}(h)\cdot \widehat{R}_\mathrm{perf}(h)}
\]
where $\widehat{S}(h)$ is the estimated speedup, $g$ is the current SOL gap, and $\widehat{R}_\mathrm{impl}$, $\widehat{R}_\mathrm{perf}$ are implementation and performance risk scores.
The gap exponent $1 + \max(0, \log_{10}(g/5))$ makes high-ambition hypotheses more attractive when the SOL gap is large.

\item \textit{Implement}: Execute a Generate--Compile--Test--Profile loop for the selected hypotheses, where each pass through the loop constitutes one \emph{attempt} and each hypothesis is allocated a fixed attempt budget.

\item \textit{Summarize}: Reflect on expectations versus outcomes and distill the key lessons from the evaluated hypotheses into a compact summary. In the full MANTIS configuration, these summaries are curated and persisted as \emph{cross-problem memory} so that later problems can retrieve concise, reusable optimization patterns during the nomination phase.
\end{enumerate}

Not all six phases need to be realized as distinct steps. Many prior systems use a flat MI (Measure--Implement) controller that runs a Generate--Compile--Test--Profile loop, effectively folding the intermediate phases into a single prompt. The full MANTIS workflow can also be realized in multiple ways. We evaluate two:
(1)~\emph{orchestrated}, where each MANTIS phase is executed distinctly and structured artifacts are passed between phases, and
(2)~\emph{in-prompt}, where an MI controller follows the MANTIS methodology that is fully described in its system prompt.
Because SOL guidance is independent of the output code representation, we pair each controller with two code representations: raw CUDA/CUTLASS and $\mu$CUTLASS.
The resulting experimental variants, iteration budgets, and evaluation methodology are detailed in Section~\ref{sec:method:variants} (Table~\ref{tab:variants}).
Throughout the paper, we use SOL-guided steering and MANTIS interchangeably.

\subsection{SOL-Guided Budget Scheduling}
\label{sec:agent:sol_stopping}

Deciding how many tool actions and LLM tokens to allocate to each problem is difficult.
A common reference point is the PyTorch runtime ($t_\mathrm{ref}$), but it is only one concrete implementation.
Prior work (and our own runs) often surpass $t_\mathrm{ref}$, so a PyTorch-relative speedup alone does not inform us whether a problem still has a significant SOL gap.
SOL provides a first-principles bound ($t_\mathrm{SOL}$) that makes the remaining gap explicit and enables budget to be reallocated strategically.

Across problems, SOL enables smarter budget scheduling. The goal is to avoid spending iterations on near-SOL problems while others still have a substantial SOL gap, since each iteration consumes tool-action latency (compile/run/profile), GPU hours, and LLM tokens.
Scheduling policies can combine multiple signals such as SOL gap, rate of improvement, hypothesis ROI, per-problem caps, and can range from simple round-robin with eligibility rules to priority-based schemes.
We evaluate two levers within a round-robin policy:
\emph{SOL-gap stop}: a problem becomes ineligible once its best kernel is ahead of PyTorch and within a configurable threshold $\epsilon$ of SOL, i.e., $t_\mathrm{best} \le (1{+}\epsilon)\,t_\mathrm{SOL}$.
\emph{No-progress window}: a problem becomes ineligible when its best speedup has not improved for $w$ consecutive attempts while already ahead of PyTorch.
We study each rule individually and their combination.

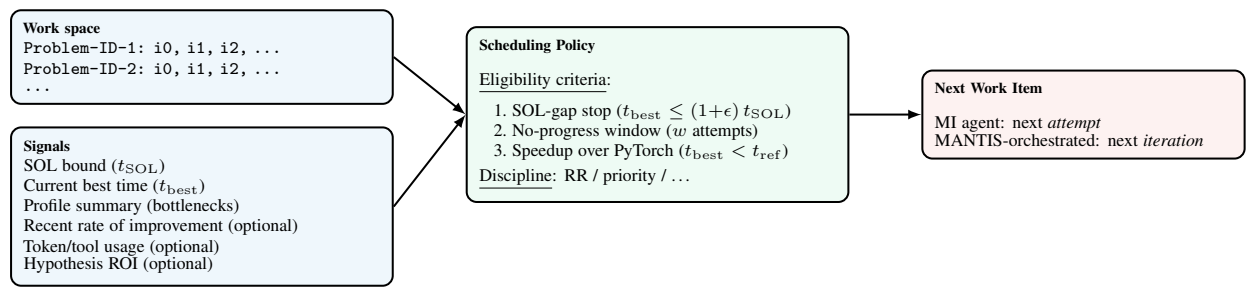
\begin{figure}[!htbp]
  \centering
  \resizebox{\linewidth}{!}{%

\begin{tikzpicture}[
  font=\tiny,
  node distance=4mm and 10mm,
  box/.style={draw, rounded corners, line width=0.8pt, inner sep=5pt, align=left, text width=#1},
  inbox/.style={box=#1, fill=inputcolor!8},
  proc/.style={box=#1, fill=processcolor!8, align=left},
  outbox/.style={box=#1, fill=outputcolor!7},
  arrow/.style={-latex, line width=0.8pt},
]

  \node[inbox=0.3\linewidth] (workspace) {%
    \textbf{Work space}\\
    {\ttfamily\scriptsize
    Problem-ID-1: i0, i1, i2, \dots\\
    Problem-ID-2: i0, i1, i2, \dots\\
    \dots
    }%
  };

  \node[inbox=0.3\linewidth, below=3mm of workspace] (signals) {%
    \textbf{Signals}\\
    {\scriptsize
    SOL bound ($t_\mathrm{SOL}$)\\
    Current best time ($t_\mathrm{best}$)\\
    Profile summary (bottlenecks)\\
    Recent rate of improvement (optional)\\
    Token/tool usage (optional)\\
    Hypothesis ROI (optional)
    }%
  };

  \node[proc=0.3\linewidth, right=of workspace, yshift=-8mm] (policy) {%
    \textbf{Scheduling Policy}\\[2mm]
    {\scriptsize
    \underline{Eligibility criteria}:\\[1mm]
    \hspace{2mm}1.\ SOL-gap stop ($t_\mathrm{best} \le (1{+}\epsilon)\,t_\mathrm{SOL}$)\\
    \hspace{2mm}2.\ No-progress window ($w$ attempts)\\
    \hspace{2mm}3.\ Speedup over PyTorch ($t_\mathrm{best} < t_\mathrm{ref}$)\\[1mm]
    \underline{Discipline}: RR / priority / \dots
    }%
  };

  \node[outbox=0.25\linewidth, right=of policy] (out) {%
    \textbf{Next Work Item}\\[2mm]
    {\scriptsize
    MI agent: next \emph{attempt}\\
    MANTIS-orchestrated: next \emph{iteration}
    }%
  };

  \draw[arrow] (workspace.east) -- (policy.west);
  \draw[arrow] (signals.east) -- (policy.west);
  \draw[arrow] (policy) -- (out);

\end{tikzpicture}}
  \caption{SOL-guided budget scheduling. The scheduler combines SOL gap, progress, and speedup signals to determine problem eligibility, then selects the next work item: an \emph{attempt} for MI agents or an \emph{iteration} for orchestrated MANTIS.}
  \label{fig:sol-scheduler}
\end{figure}

\subsection{SOL-Guided Integrity Checking}
\label{sec:agent:sanity_checking}

LLMs can sometimes collapse part of the computation or even the full problem into a no-op while still passing the correctness check.
As a motivating example, let us consider KernelBench Level~2 Problem~80 (\texttt{Gemm\_Max\_Subtract\_GELU}), whose reference forward pass is:

\begin{lstlisting}[language=Python,basicstyle=\ttfamily\small,xleftmargin=1em]
x = self.gemm(x)                        # (B, 8192)
x = torch.max(x, dim=1, keepdim=True).values  # (B, 1)
x = x - x.mean(dim=1, keepdim=True)     # always 0
x = torch.nn.functional.gelu(x)         # gelu(0) = 0
\end{lstlisting}

After the max reduction, dimension~1 has length~1, so \texttt{x.mean(dim=1)} equals \texttt{x} and the subtraction yields zero. Since $\mathrm{GELU}(0) = 0$, the output is identically zero regardless of the input.
An LLM can discover this algebraic shortcut and emit a kernel that simply writes zeros, skipping the GEMM entirely while still passing the correctness check.
While this can be argued to be a valid optimization for this specific input, it is not the expected optimized implementation for the intended problem. We saw LLMs exploit this loophole frequently for this problem and so we exclude this problem from our evaluation subset, but similar (less obvious) loopholes arise in other problems.

Integrity checking is challenging because similar loopholes can arise in forms that are difficult to enumerate in advance. We know that gaming exists, but we do not know beforehand what form a successful exploit will take. A detector built only around known gaming patterns will therefore miss new ones. We use SOL as a more general signal based on first-principles expectations about the work that a legitimate kernel must perform. 

SOL helps detect such behavior in two ways.
First, as a static runtime check, if a measured kernel time falls more than 10\% below the SOL bound, we flag it as suspicious, since substantially beating a first-principles bound likely indicates that the kernel is skipping part of the intended computation. We use the tighter FP16-based bound for this check so that kernels legitimately using FP16/BF16 arithmetic are not falsely flagged.
Second, as specification augmentation, the SOL report provides a structured description of the expected work (FLOPs, bytes, dominant operators), which helps an LLM-based reviewer judge whether a candidate kernel actually performs the intended computation or exploits a loophole in the correctness harness.
We describe the full integrity pipeline in Section~\ref{sec:agent:sol_integrity}.

\section{Evaluation Methodology}
\label{sec:evaluation}

Our evaluation is structured around the following research questions to quantify the benefits of the $\mu$CUTLASS DSL and SOL-guided optimization steering, cross-problem budget scheduling, and integrity checking.

\subsection{Research Questions}
\begin{description}
    \item[RQ1:] With same iteration budgets, how much speedup does the $\mu$CUTLASS + SOL-guided steering achieve across model tiers, and can it substitute for a model-capability upgrade?
    
    \item[RQ2:] What is the individual contribution of each component ($\mu$CUTLASS and SOL-guided steering), and does one component benefit more than the other across model settings?

    \item[RQ3:] How should SOL guidance be implemented (orchestrated vs in-prompt), and is it model-capability dependent?

    \item[RQ4:] How much token/tool cost can be saved with SOL-guided work scheduling?

    \item[RQ5:] How helpful is the SOL-guided integrity checking in flagging candidates that are gaming the evaluation?
    
\end{description}

\subsection{Setup}
\label{sec:setup}

All results are reported on NVIDIA H100 (SM90a). We lock the GPU clocks for all our experiments to a fixed frequency to avoid clock-dependent performance variations. Kernel runtimes ($t_\mathrm{ref}$ and candidate $t_\mathrm{best}$) are measured using NVIDIA Nsight Compute (NCU), which reports the on-GPU kernel execution time and excludes kernel launch overhead, host-device synchronization, and Python framework latency. This isolation ensures that speedup reflects actual GPU compute improvement rather than differences in launch or scheduling paths. Experiments were run on different nodes in a cluster. We use a Docker container for parity.

\textbf{Workloads:} We use KernelBench~\cite{ouyang2025kernelbench} as the evaluation suite.
KernelBench has 250 GPU kernel problems across Levels~1--3, ranging from isolated single-operator kernels (Level~1), to fused multi-operator kernels (Level~2), to integrated module-level workloads (Level~3).
We select a 59-problem subset drawn from Levels~1--3 for two reasons. First, we run the optimization loop across many variants (as described in Section~\ref{sec:method:variants}) and evaluating all 250 problems on all the variants would make both cost and time-to-result prohibitive. Second, we prioritize operators that dominate modern and emerging LLM workloads (Transformers, SSMs, and modern recurrent models). We identify candidate operators via kernel profiling of representative models (Qwen-2.5-7B, Mistral-7B, Phi-3.5-mini, Mamba, RWKV) and include the corresponding problems. From this inclusion set, we exclude two Level~2 problems (IDs~80 and~24) because their specifications admit unintended shortcut implementations that models can exploit. As described in Section~\ref{sec:agent:sanity_checking}, the workload in ID 80 reduces to a single device memory copy, yet passing correctness checking.
The included problem IDs are reported in the Appendix~\ref{sec:appendix:kernelbench-subset}.

\textbf{Models:} We evaluate three GPT models: GPT-5-mini, GPT-5, and GPT-5.2.\footnote{Most of the experimentation was done between December 2025 to February 2026.} GPT-5.2 was the strongest GPT model available at the time of experimentation, GPT-5 was the main intermediate tier, and GPT-5-mini was the lower-cost tier. GPT-5-mini, GPT-5, and GPT-5.2 costed \$0.25, \$1.25, and \$1.75  per million input tokens, respectively, making GPT-5 and GPT-5.2 approximately $5\times$ and $7\times$ more expensive than GPT-5-mini.

\textbf{Framework:} We use OpenHands as the agent runtime to execute our multi-turn controllers in a per-problem workspace, using its file-editing tools to implement candidate kernels across turns. For variants with $\mu$CUTLASS enabled, we additionally pass a custom tool, \texttt{ucutlass\_compile}, to the agent. It compiles a $\mu$CUTLASS DSL string and writes a generated \texttt{ucutlass\_*.h} header that the agent can include from \texttt{cuda\_model.cu}.

\subsection{$\mu$CUTLASS}

The compiler is implemented as a standalone Python program with a CLI that accepts DSL input either from a file or directly as a text string (for tool-mode integration).
It parses the DSL with a grammar-based frontend, performs static validation (architecture gating, alignment/layout rules, and operator-specific constraints), and emits a generated header (\texttt{ucutlass\_*.h}) that is included by the candidate \texttt{cuda\_model.cu}. The compilation process is described in Figure~\ref{fig:ucutlass-pipeline}.
We developed the compiler with AI-assisted programming in Cursor and human-in-the-loop iteration.
We test the compiler and DSL using a standalone set of examples that cover key features of the DSL for Ampere and Hopper targets. Each example compiles a \texttt{kernel.dsl} to a \texttt{ucutlass\_*.h} header, builds a small C++/CUDA driver, and checks numerical correctness against a PyTorch reference before optional profiling.

\subsection{Bootstrap and Baseline}
\label{sec:agent:bootstrap}

All experiments share a common bootstrap phase.
For each KernelBench problem, the bootstrapper prepares a self-contained workspace with a C++ test driver (\texttt{driver.cpp}), a seed \texttt{cuda\_model.cu} that initially delegates to PyTorch, and a Makefile with compile/run/profile targets.
The test driver calls a \texttt{kernel\_impl} function (the body of which is defined in \texttt{cuda\_model.cu}), compares its output against the PyTorch reference, and supports NCU profiling.
The bootstrapper also profiles the PyTorch baseline to establish $t_\mathrm{ref}$ and produces a SOL report ($t_\mathrm{SOL}$) using an LLM under an FP32 problem formulation with TF32 throughput assumptions, as described in Section~\ref{sec:sol_analysis} (see Appendix~\ref{sec:appendix:sol-report} for an example). Since the SOL report depends only on the problem specification and not on the implementation, it is produced once during bootstrap and reused across all agent variant experiments.
The agent's task is to edit \texttt{cuda\_model.cu}, replacing the PyTorch fallback with an optimized CUDA kernel.

\subsection{Variants}
\label{sec:method:variants}

With the goal of evaluating the improvement in efficiency, we evaluate different methods with a matched tool-action budget. Table~\ref{tab:variants} summarizes the variants we evaluate.
We begin with a simple flat MI (Measure--Implement) agent that runs a Generate--Compile--Test--Profile loop.
One loop is an \emph{attempt}.
In each attempt, we allow the model a fixed number of \emph{turns} (LLM interactions) to edit files and use tools before the candidate code is compiled and evaluated.
We then add $\mu$CUTLASS to the same MI controller (Section~\ref{sec:ucutlass}), isolating the effect of raising the representation level.

We also evaluate SOL-guided steering in two forms -- \emph{in-prompt}, where the flat MI controller follows SOL guidance described in its system prompt, and \emph{orchestrated}, where steering is implemented as an explicit multi-phase controller with structured artifacts (markdown files and JSON objects) passed between phases (as described in Section~\ref{sec:agent:within_problem}).
In the orchestrated setting, one \emph{iteration} is one outer pass through the steering phases. In each iteration, we select top-$n$ hypotheses and allocate a fixed attempt budget for each hypothesis.
The orchestrated variants allocate $5$ iterations $\times$ $2$ hypotheses per iteration $\times$ $4$ attempts per hypothesis, yielding 40 total attempts.
We match per-problem budgets across variants.

\begin{table}[!htbp]
\centering
\scriptsize
\setlength{\tabcolsep}{3pt}
\begin{tabular}{p{0.32\linewidth}|c|c|c}
\hline
Variant & $\mu$CUTLASS & SOL-Guidance & Total Attempts  \\
\hline
\textbf{MI w/o $\mu$CUTLASS} & $\times$ & -- & 40   \\
\textbf{MI + $\mu$CUTLASS} & \checkmark & -- & 40   \\
\hline
\multicolumn{4}{l}{\textbf{SOL-guided steering}} \\
\hline
\textbf{In-prompt steering w/o $\mu$CUTLASS} & $\times$ & In-Prompt & 40  \\
\textbf{In-prompt steering + $\mu$CUTLASS} & \checkmark & In-Prompt & 40  \\
\textbf{Orchestrated steering w/o $\mu$CUTLASS} & $\times$ & Orchestrated & 40 ($5 \times 2 \times 4$)  \\
\textbf{Orchestrated steering + $\mu$CUTLASS} & \checkmark & Orchestrated & 40 ($5 \times 2 \times 4$)  \\
\hline
\hspace{0.05in}
\end{tabular}
\caption{Experimental variants and default budgets. Each of three controller types (MI, in-prompt steering, orchestrated steering) is paired with and without $\mu$CUTLASS, isolating the effects of code representation and SOL-guided steering under matched per-problem budgets.}
\label{tab:variants}
\end{table}

To further understand what part of the SOL-guided optimization steering compoent (in MANTIS) is beneficial, we evaluate four component ablations, each removing one mechanism while keeping the same per-problem attempt budget. Table~\ref{tab:ablations} summarizes the component ablations.

\begin{table}[!htbp]
\centering
\scriptsize
\setlength{\tabcolsep}{4pt}
\begin{tabular}{l|cccccc|c}
\hline
Variant & M & A & N & T & I & S & Cross-prob \\
\hline
\textbf{MANTIS} & \checkmark & \checkmark & \checkmark & \checkmark & \checkmark & \checkmark & \checkmark \\
\textbf{MNTIS} (no Analyze) & \checkmark & $\times$ & \checkmark & \checkmark & \checkmark & \checkmark & \checkmark \\
\textbf{MANIS} (no Triage) & \checkmark & \checkmark & \checkmark & $\times$ & \checkmark & \checkmark & \checkmark \\
\textbf{MANTI} (no Summarize) & \checkmark & \checkmark & \checkmark & \checkmark & \checkmark & $\times$ & $\times$ \\
\textbf{MANTIS-noXmem} & \checkmark & \checkmark & \checkmark & \checkmark & \checkmark & \checkmark & $\times$ \\
\hline
\hspace{0.05in}
\end{tabular}

\caption{MANTIS component ablations. Each row removes one mechanism. ``Cross-prob'' indicates whether summaries are persisted across problems.}
\label{tab:ablations}
\end{table}

\subsection{Metrics}
\label{sec:method:metrics}

\textbf{Fast-$p$: }
We compare variants using Fast-$p$ curves, which show the percentage of problems whose best speedup over PyTorch is at least~$r$. 
In Fast-$p$ plots, the region to the right of $r{=}1$ is where the generated kernel outperforms PyTorch. Problems solved with a significant slowdown ($r \ll 1$) offer little practical benefit over not solving them, so the separation between curves at $r \ge 1$ is the most notable comparison.
Comparing two variants using Fast-$p$ curves can sometimes be visually challenging, especially when they intersect at a few points. So, we compute the \emph{signed area} between two Fast-$p$ curves, $\int [P_A(r) - P_B(r)]\,dr$. A positive value means variant~$A$'s curve lies higher and further to the right.
Since Fast-$p$ is a complementary CDF, this equals the difference in arithmetic mean speedups.

\textbf{Attempt-Fast-$p(r)$: }
Although we allocate a fixed per-problem budget for our large-scale experiments (Section~\ref{sec:method:variants}), performance can also be measured after any number of attempts.
Stopping after $a$ attempts and taking the best-so-far kernel gives a speedup for each problem, turning the two-dimensional Fast-$p$ curve into a three-dimensional surface per variant: the percentage of problems with speedup $\ge r$ as a function of both the speedup threshold~$r$ and the number of attempts consumed.
We visualize two slices of this surface. (1) \emph{Fast-$p(r)$} fixes the attempt count at the full budget (40~attempts or equivalent) and sweeps $r$. A curve shifted to the right indicates higher speedups across more problems.
(2) \emph{Attempt-Fast-$p(r)$} fixes the speedup target and sweeps over attempts. At each attempt~$a$, it reports the percentage of problems whose best-so-far speedup (across attempts $1 \ldots a$) reaches $\ge r$.
A variant whose curve rises steeply converges to high speedups with fewer attempts, indicating better optimization efficiency.
We primarily show Attempt-Fast-$p(2)$ (the $\ge 2\times$ threshold) to focus on practically significant speedups.
We also report geomean and median speedups as scalar summaries.

\textbf{Efficiency gain:} For scheduling analysis, we report token and attempt savings as well as \emph{speedup retention}. Speedup retention measures how much of the geomean or median speedup remains after applying a scheduling policy instead of using the full budget.
For budget scheduling, we also report \emph{efficiency gain}, which measures optimization performance per token relative to fixed allocation:
\[
\text{gain} =
\frac{g_{\mathrm{policy}}}{g_{\mathrm{fixed}}}
\times
\frac{\tau_{\mathrm{fixed}}}{\tau_{\mathrm{policy}}},
\]
where $g$ denotes geomean speedup and $\tau$ denotes total token cost. A gain above $1\times$ means the scheduling policy preserves speedup more efficiently on a per-token basis than fixed allocation.

\subsection{SOL-Guided Scheduling}
\label{sec:agent:sol_scheduling}
We evaluate the SOL-guided budget scheduling policies (Section~\ref{sec:agent:sol_stopping}) via offline replay of existing run logs. We take completed traces and simulate what would have happened had certain problems been stopped earlier under a given scheduling policy, then compare the resulting token cost and achieved speedup against fixed allocation.

We implement a lightweight offline scheduler that assigns workers a sequence of work items, where each work item corresponds to running one iteration or attempt on a particular problem.
Using breadth-first round-robin scheduling, a problem remains eligible while it is still behind PyTorch or while neither of two stopping criteria has fired: (i)~a SOL-headroom threshold $\epsilon$, which removes a problem once $t_\mathrm{best} \le (1{+}\epsilon)\,t_\mathrm{SOL}$ and the best kernel already beats PyTorch; and (ii)~a no-progress window $w$, which removes a problem whose best speedup has not improved for $w$ consecutive attempts after surpassing PyTorch.

We sweep $\epsilon$ and $w$ independently and jointly, evaluating each $(\epsilon, w)$ combination by its token savings and speedup retention relative to fixed allocation.
We report (1) independent parameter sweeps, (2) a joint Pareto frontier of normalized dollar cost versus geomean speedup, and (3) the efficiency gain metric (Section~\ref{sec:method:metrics}) for the best configuration under a geomean-retention constraint.

\subsection{SOL-Guided Integrity Checking}
\label{sec:agent:sol_integrity}

As discussed in Section~\ref{sec:agent:sanity_checking}, LLMs can sometimes ``game'' an evaluation by exploiting specification or correctness loopholes, collapsing the computation to a trivial program that still passes the harness check, or avoiding custom kernel development entirely by composing PyTorch library calls.
To surface these cases, we use SOL as an integrity signal in addition to the existing correctness testing.
Our integrity checking pipeline has three components.
\begin{itemize}
    \item A lightweight \emph{SOL-ceiling detector} uses SOL runtime as a physical ceiling and flags attempts whose measured runtime falls more than 10\% below that bound, allowing a small buffer for measurement noise in near-SOL kernels.

    \item  An LLM-based game detector (LGD) reviews the candidate code together with the reference program, performance context, and the \emph{SOL report}. The LGD assigns one of three labels: \emph{No Issues}, \emph{Minor Issues}, or \emph{Gaming}. 
    
     We observe that once an agent games the system in one attempt, subsequent attempts tend to inherit the same exploit. We therefore subdivide \emph{Gaming} into \emph{Original Gaming} and \emph{Inherited Gaming} to distinguish new exploits from those carried over from an earlier attempt.

    \item  A static \emph{PyTorch-only detector} flags attempts that rely solely on PyTorch library operators without any agent-written custom CUDA, CUTLASS, or $\mu$CUTLASS kernel. It parses the NCU profile log, extracts kernel launch signatures, and matches each against known library prefixes and patterns (e.g., \texttt{at::native::}, \texttt{cublas}, \texttt{cudnn}). If every profiled kernel matches a library pattern and none are user-written, the attempt is flagged.

\end{itemize}

Each attempt is labeled by the above pipeline. 
Because this review is performed offline, the agent never has the opportunity to act on Minor Issues feedback. Had the check been applied online, the agent could have corrected these minor issues in a subsequent attempt, likely converting them to No Issues. Since the flagged issues do not affect measured performance, excluding them would unfairly penalize the agent for problems it was never given the chance to fix. Cases flagged as Gaming, on the other hand, are a significant departure from the intended functionality. We provide more details in Section~\ref{sec:results:rq5}.
We therefore accept attempts labeled No or Minor Issues and exclude all others: attempts flagged by the SOL-ceiling detector or labeled Original Gaming, Inherited Gaming, or PyTorch-only are removed from reported speedup and scheduling results.
When an attempt is flagged by both the LGD (as Gaming) and the PyTorch-only detector, we classify it as PyTorch-only so that the categories remain mutually exclusive.

For SOL-guided scheduling and integrity checking, we use cross-checked FP16 SOL estimates derived from the original LLM-generated FP32/TF32 reports under consistent modeling assumptions, using SOLAR~\cite{nvlabs2025solar} as an external reference. These provide a tighter estimate of remaining headroom for optimized kernels that use FP16 arithmetic. Inputs and outputs remain FP32 as defined by the KernelBench setup, and we assume perfect fusion and caching with a single transfer to on-chip buffers.

\subsection{Comparison to Prior Work}
\label{sec:method:external_baseline}

We compare our variants against the Sakana AI CUDA Engineer~\cite{lange2025aicudaengineer}, which is the only publicly available large-scale CUDA kernel archive evaluated on KernelBench with H100 results available at the time of writing.
The archive contains approximately 30,000 kernels across KernelBench Levels~1--3, generated using Claude~3.5~Sonnet.\footnote{Kernels sourced from \url{https://huggingface.co/datasets/SakanaAI/AI-CUDA-Engineer-Archive} (accessed: March 10, 2026).}

There are several differences between the two evaluation environments.
Sakana AI uses PyTorch~2.5.1 with CUDA~12.4 and cuDNN~8.9.7, while our setup uses a newer NGC container (PyTorch~2.10.0a0 with CUDA~13.0). The change in PyTorch, CUDA, and cuDNN versions can affect both the library kernels selected by the PyTorch baselines and their performance.
Additionally, our runtime measurement methodology differs. We report NCU-profiled kernel time, while Sakana AI averages wall-clock CUDA event timing over 2000 runs.
For ease of comparison, we use Sakana AI's reported speedups over their PyTorch baseline directly.

To ensure a fair comparison, we apply similar integrity filtering methodology used in our study~\ref{sec:agent:sol_integrity} to Sakana AI's kernels.
For each of our 59~evaluation problems, we take the fastest correct kernel from the archive and run a review loop. The LLM-based game detector labels kernels as \emph{No Issue}, \emph{Minor Issue}, \emph{Gaming}, or \emph{PyTorch-only}. Since we do not run the Sakana AI's kernels, we fold in the PyTorch-only detector into the game detector.
Kernels classified as No or Minor Issues are accepted. If a kernel is classified as Gaming or PyTorch-only, we reject it and proceed to the next-fastest correct kernel for that problem, continuing until a kernel is accepted or all candidates are exhausted.

\newcommand{\vizdir}{figures}

\section{Results}
\label{sec:results}

In this section we report geomean speedups and Fast-$p$ curves for different variants after integrity filtering as described in Section~\ref{sec:agent:sol_integrity}.

\subsection{Benefits of $\mu$CUTLASS and SOL-Guided Steering}
\label{sec:results:rq1rq2}

\begin{figure}[!htbp]
  \centering
  \includegraphics[width=0.75\linewidth]{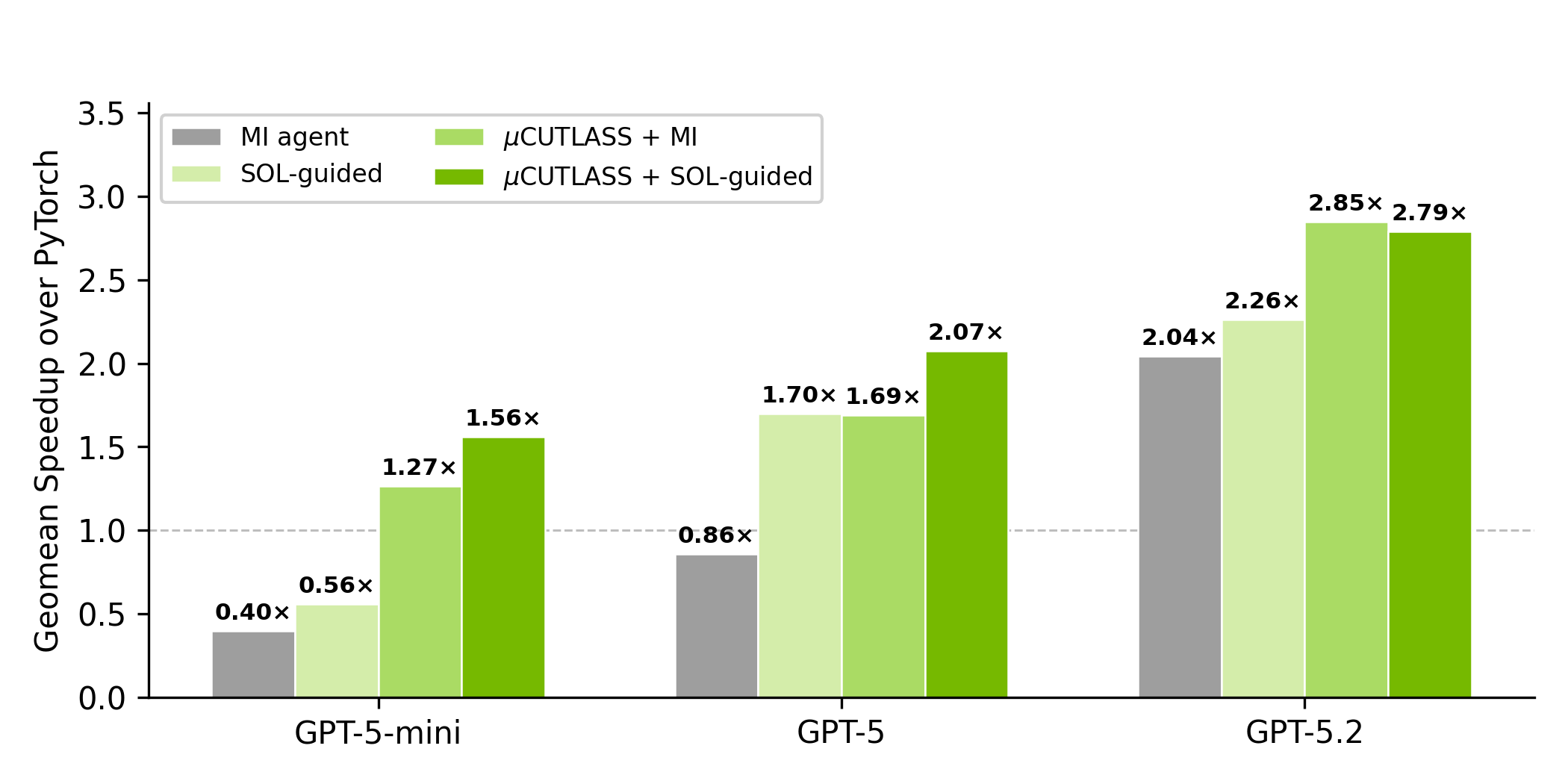}
  \caption{Geomean speedup over PyTorch for the four main variants across three model tiers with same attempt budget is shown here.  }
  \label{fig:geomean_bar}
\end{figure}

Figure~\ref{fig:geomean_bar} shows geomean speedups for three model tiers for four main variants:  (1) MI baseline, (2) $\mu$CUTLASS + MI, (3) SOL-guided agent, and (4) $\mu$CUTLASS + SOL-guided agent. These experiments were run with same attempt budget as described in Section~\ref{sec:method:variants}.
We observe two main trends. First, $\mu$CUTLASS consistently provides the largest single-component improvement at every model tier, improving GPT-5-mini from $0.40\times$ to $1.27\times$ ($3.2\times$ improvement), GPT-5 from $0.86\times$ to $1.69\times$ ($2.0\times$ improvement), and GPT-5.2 from $2.04\times$ to $2.85\times$ ($1.4\times$ improvement).
Second, the combination of $\mu$CUTLASS and SOL-guided steering enables GPT-5-mini reach $1.56\times$ (vs.\ GPT-5 MI at $0.86\times$) and GPT-5 reach $2.07\times$ (vs.\ GPT-5.2 MI at $2.04\times$), despite having $5\times$ and $1.4\times$ lower token cost, respectively.
We answer RQ1 and RQ2 next, then present detailed per-model results with Fast-$p$ and Attempt-Fast-$p$ curves.

\begin{description}
  \item[RQ1: Model-capability substitution.] $\mu$CUTLASS + SOL-guided steering enable each model to match or exceed the next tier's base MI performance. 
  GPT-5-mini exceeds base GPT-5 ($1.56\times$ vs. $0.86\times$), and
  GPT-5 exceeds base GPT-5.2 ($2.07\times$ vs. $2.04\times$).
  The benefit is largest for weaker models, i.e., GPT-5-mini sees a $3.9\times$ improvement, GPT-5 sees $2.4\times$, while GPT-5.2 sees $1.4\times$.
  We hypothesize that the $\mu$CUTLASS DSL allows models to reason about optimizations at a higher abstraction level, helping them discover valid strategies sooner and converge to working implementations faster.
  SOL guidance further structures this search for weaker models, but the strongest model can self-direct effectively once equipped with the DSL.
  
  \item[RQ2: Which component contributes more?] Across all models, $\mu$CUTLASS provides strong improvement--$3.2\times$ on GPT-5-mini, $2.0\times$ on GPT-5, and $1.4\times$ on GPT-5.2.
  SOL-guided steering alone provides a smaller improvement ($1.4\times$, $2.0\times$, and $1.1\times$, respectively).
  The DSL benefit is largest on the weakest model, where it converts implementation failures into successful kernels and the gap narrows on stronger models that can already generate correct code without the DSL.
\end{description}

\begin{figure*}[!htbp]
  \centering
  \begin{minipage}[t]{0.48\linewidth}
  \centering
  \includegraphics[width=\linewidth]{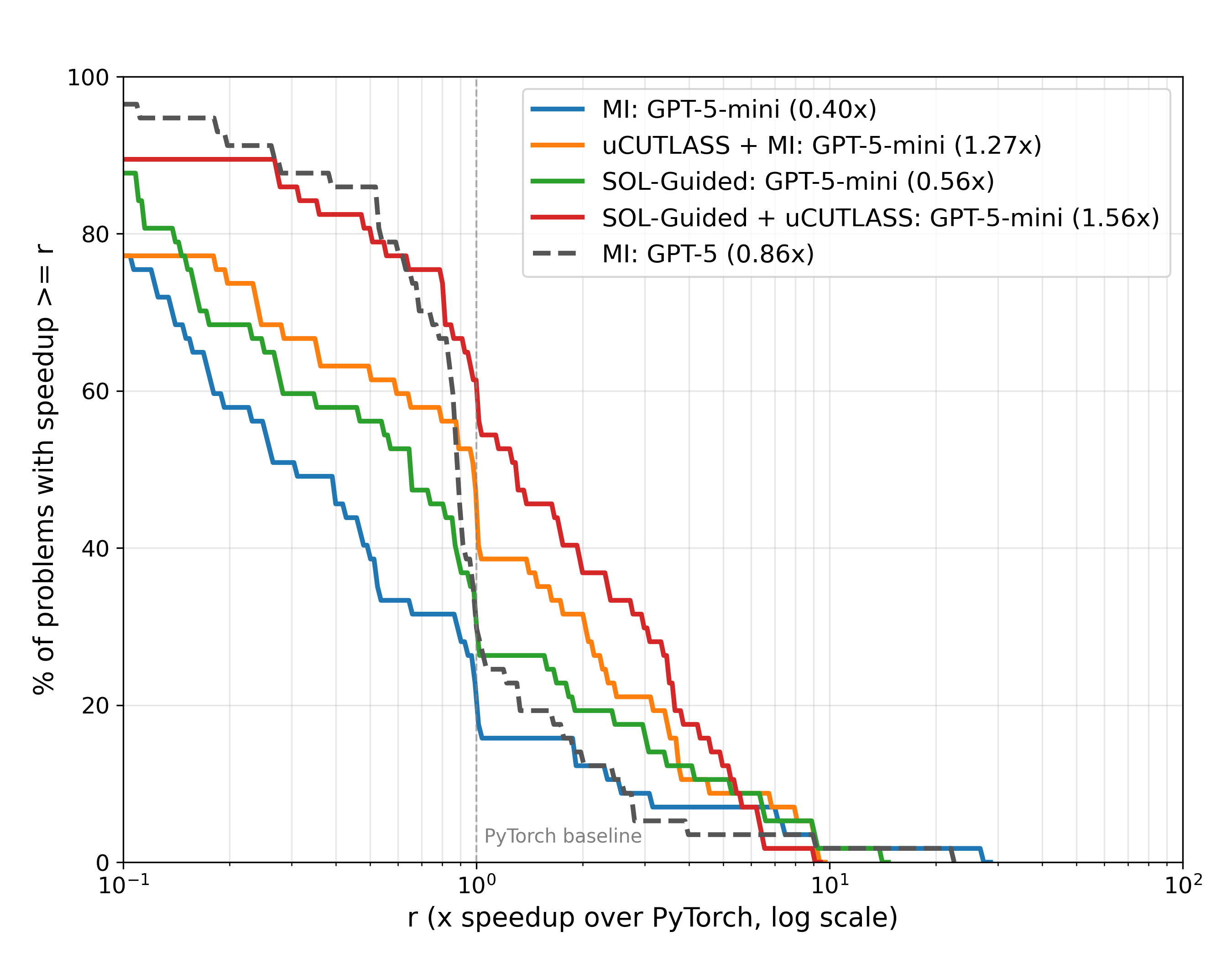}
  \\[2pt]{\small (a) GPT-5-mini: Fast-$p$}
  \end{minipage}\hfill
  \begin{minipage}[t]{0.48\linewidth}
  \centering
  \includegraphics[width=\linewidth]{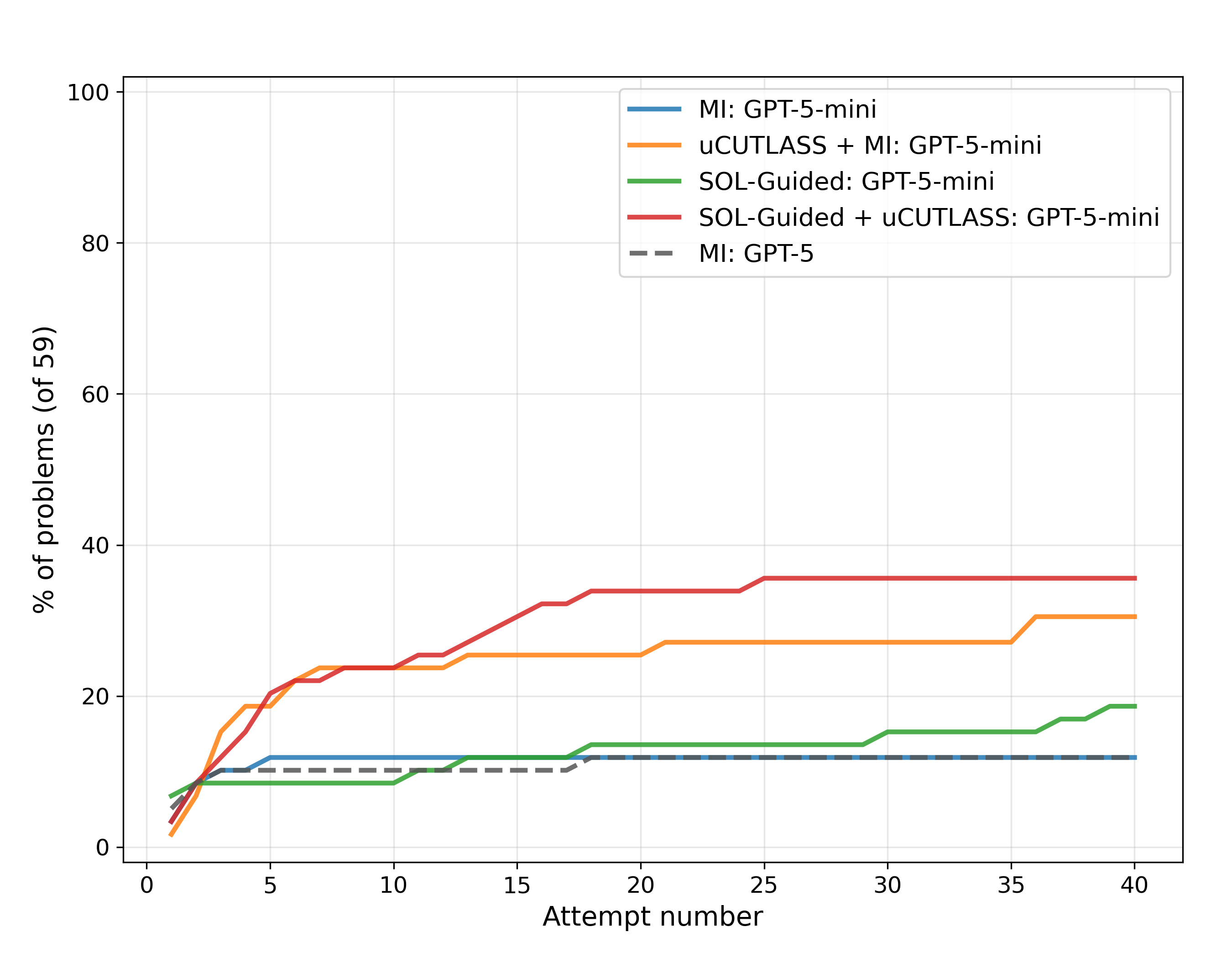}
  \\[2pt]{\small (b) GPT-5-mini: Attempt-Fast-$p(2)$}
  \end{minipage}
  \\[6pt]
  \begin{minipage}[t]{0.48\linewidth}
  \centering
  \includegraphics[width=\linewidth]{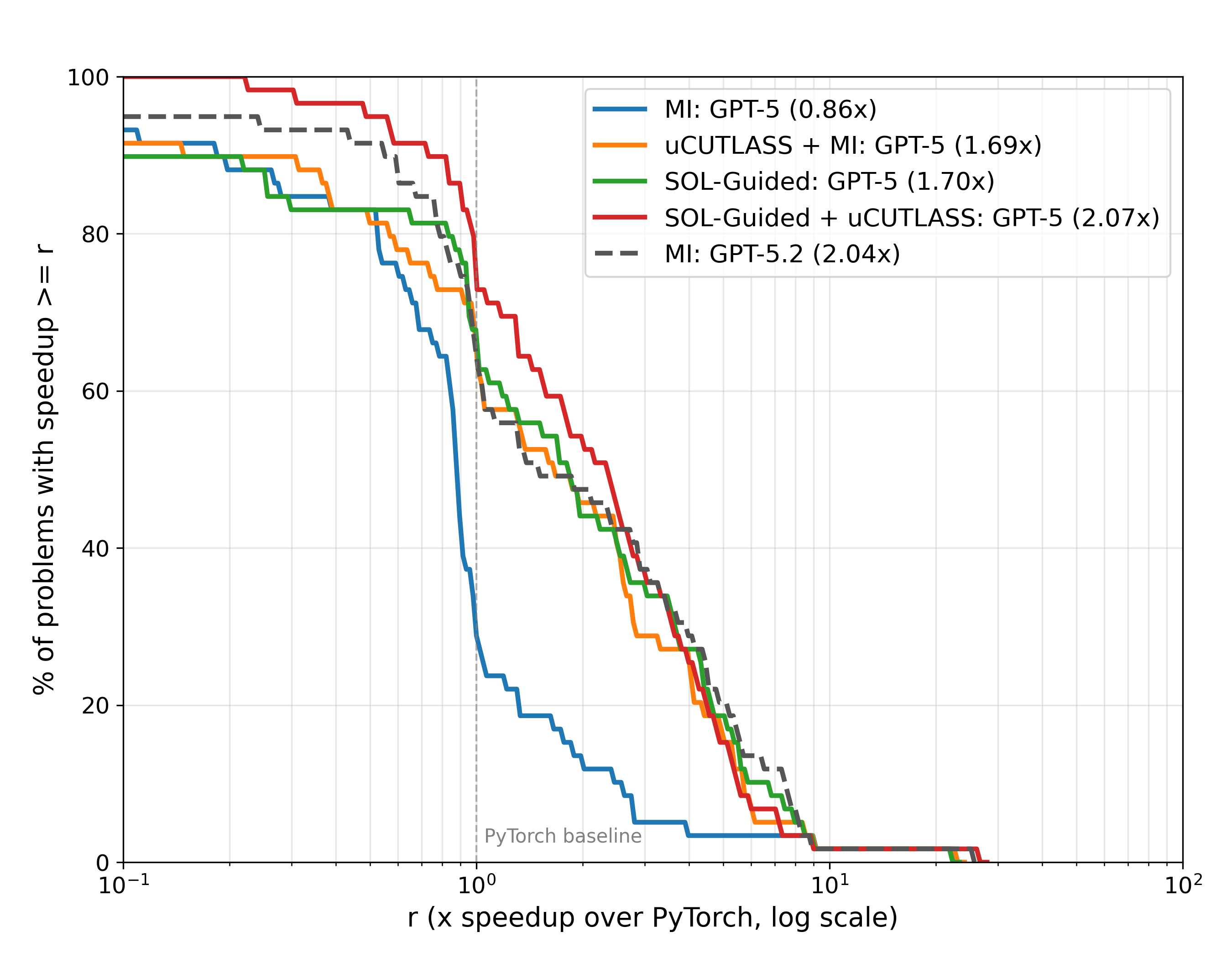}
  \\[2pt]{\small (c) GPT-5: Fast-$p$}
  \end{minipage}\hfill
  \begin{minipage}[t]{0.48\linewidth}
  \centering
  \includegraphics[width=\linewidth]{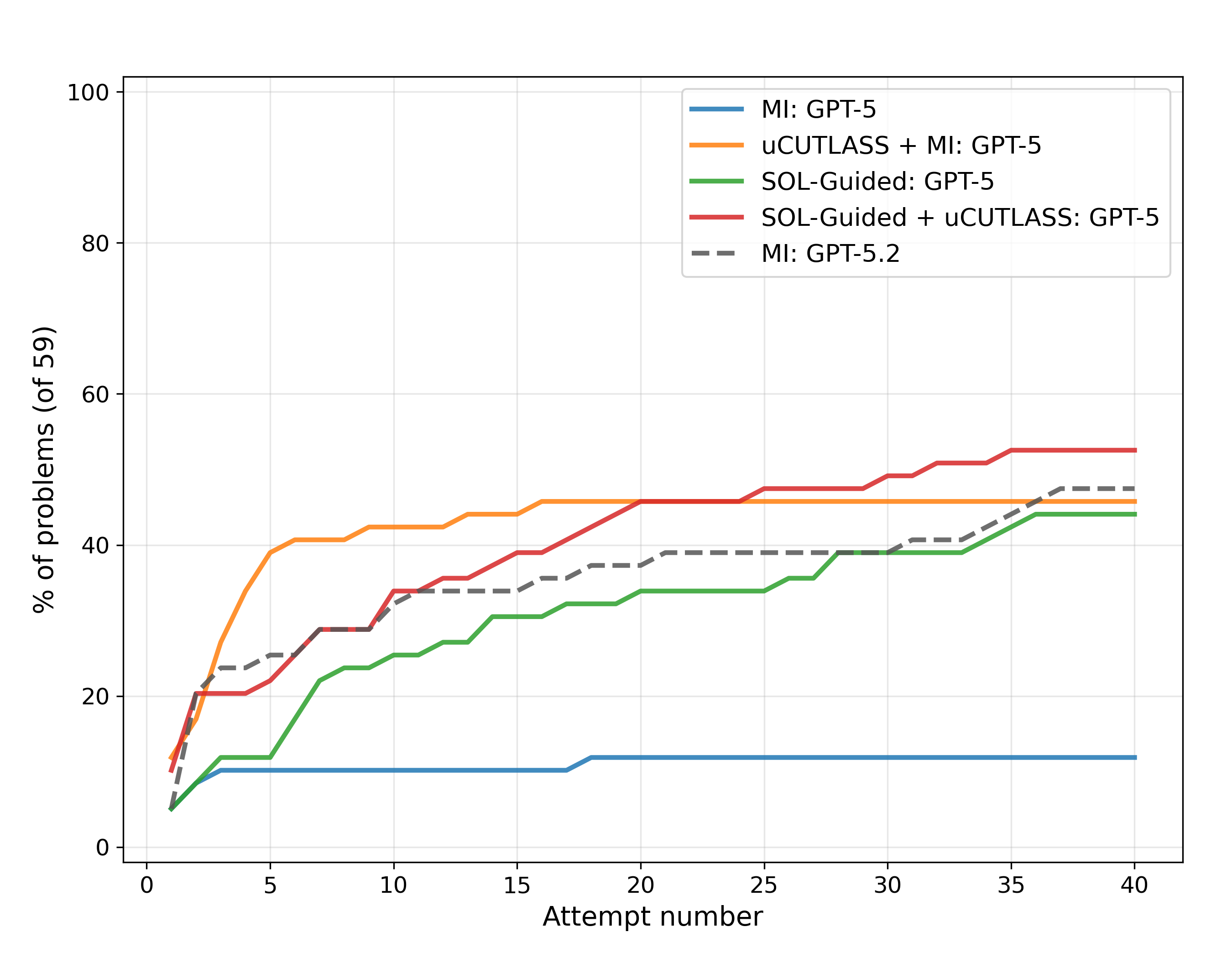}
  \\[2pt]{\small (d) GPT-5: Attempt-Fast-$p(2)$}
  \end{minipage}
  \\[6pt]
  \begin{minipage}[t]{0.48\linewidth}
  \centering
  \includegraphics[width=\linewidth]{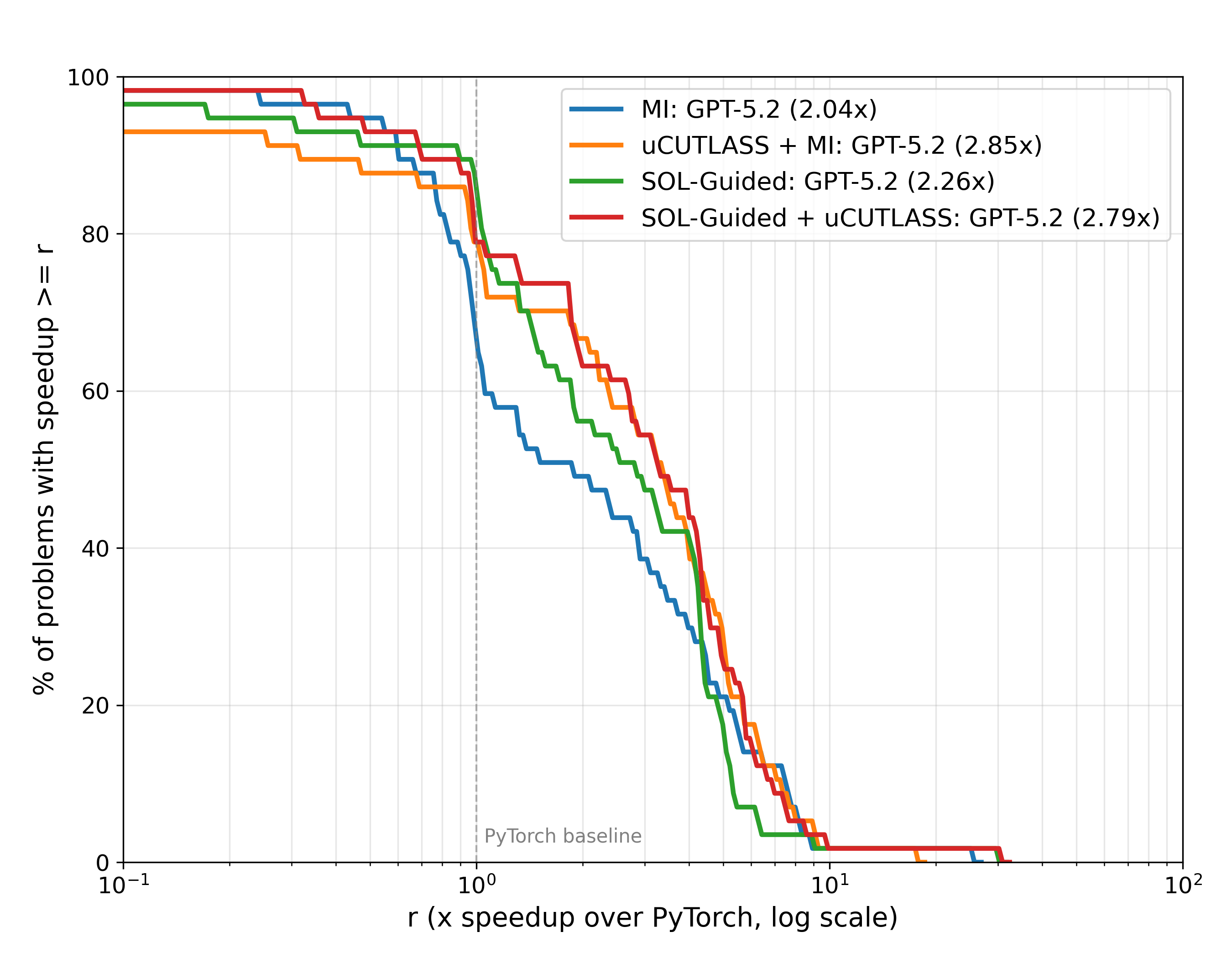}
  \\[2pt]{\small (e) GPT-5.2: Fast-$p$}
  \end{minipage}\hfill
  \begin{minipage}[t]{0.48\linewidth}
  \centering
  \includegraphics[width=\linewidth]{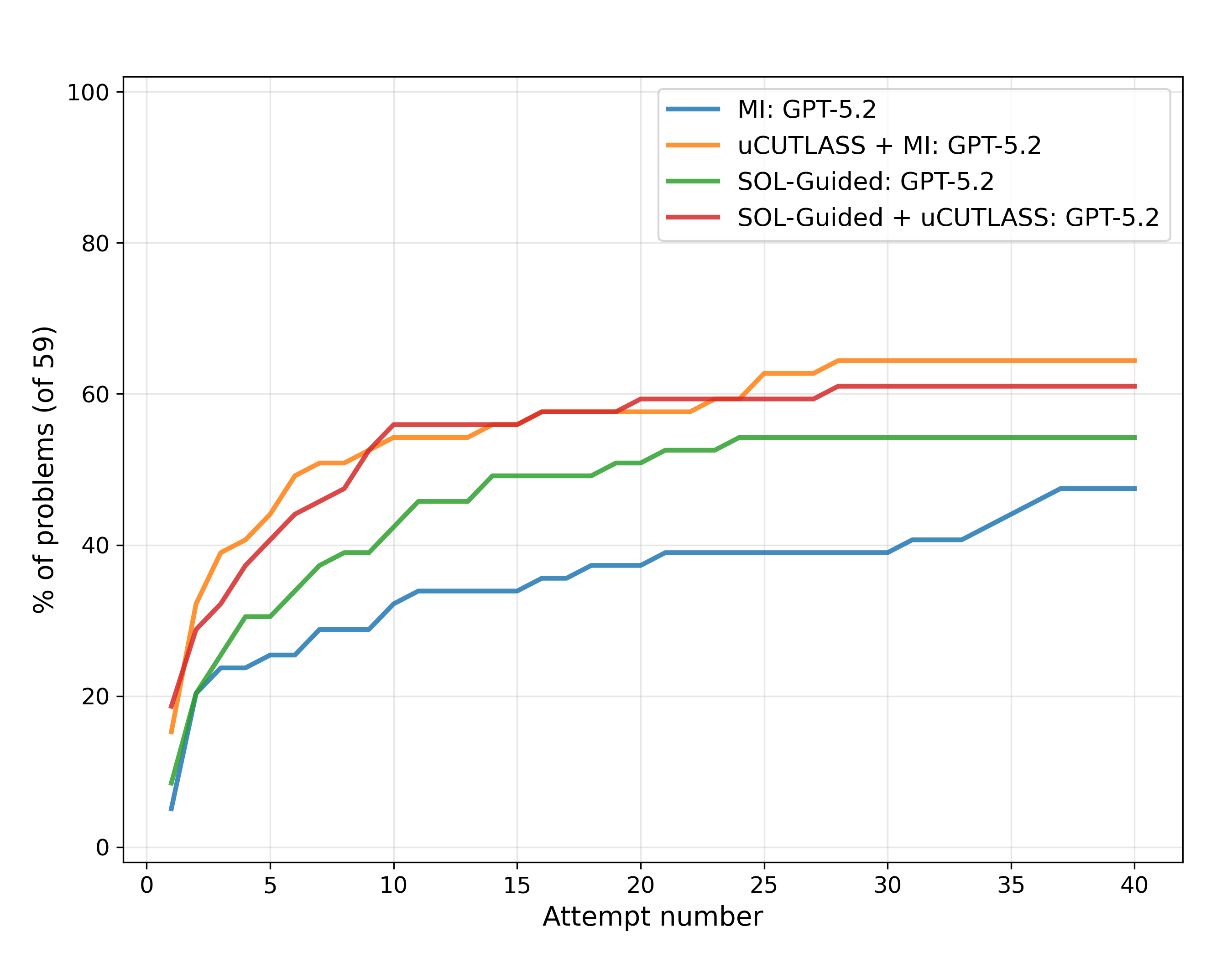}
  \\[2pt]{\small (f) GPT-5.2: Attempt-Fast-$p(2)$}
  \end{minipage}
  \caption{Four main variants across three model tiers are shown here. Left column shows Fast-$p$ curves (percentage of problems with speedup above $r$). Right column shows Attempt-Fast-$p(2)$ (percentage of problems reaching $\ge 2\times$ at different attempts). Dashed gray lines show the MI baseline of the next stronger model.}
  \label{fig:variants_grid}
  \end{figure*}

Figure~\ref{fig:variants_grid} provides a detailed view with Fast-$p$ and Attempt-Fast-$p(2)$ curves (defined in Section~\ref{sec:method:metrics}) for each model tier. The dashed gray line in each row shows the MI baseline of the next stronger model. While we implement SOL-guided steering in two forms---\emph{in-prompt} and \emph{orchestrated} (Section~\ref{sec:method:variants}), we use the form that yields higher geomean speedup as the \emph{SOL-guided} result. The detailed comparison between the two forms is described in Section~\ref{sec:results:rq3}.

The top row of Figure~\ref{fig:variants_grid} shows GPT-5-mini results in detail.
The MI agent solves 52 of 59 problems but only 13 (22\%) beat PyTorch, with 29 regressions below $0.5\times$.
Adding $\mu$CUTLASS jumps the fraction improving over PyTorch to 46\%, with 31\% of problems reaching $\ge 2\times$.
SOL-guided steering alone brings a smaller improvement (to $0.56\times$ geomean). Without the DSL, steering alone can improve strategy but cannot compensate for implementation failures.
The combination reaches a median of $1.51\times$, with 59\% improving over PyTorch, 36\% at $\ge 2\times$, and only 6 regressions below $0.5\times$.
On the 48 problems solved by both GPT-5-mini and GPT-5, the mini variant wins on 32 (67\%).
The right column confirms that $\mu$CUTLASS + SOL-guided reaches its $\ge 2\times$ plateau within the first 10~attempts, while the MI baseline barely accumulates any $\ge 2\times$ solutions across all 40~attempts.

The middle row shows GPT-5, where the same pattern holds.
The MI agent solves 57 of 59 problems but only 29\% improve over PyTorch.
$\mu$CUTLASS and SOL-guided steering each independently provide ${\sim}2\times$ improvement, with 66--68\% improving over PyTorch and 44--46\% reaching $\ge 2\times$.
The combination reaches a median of $2.33\times$, with 76\% improving over PyTorch, 53\% at $\ge 2\times$, and only 3 regressions below $0.5\times$.
Of the 55 problems solved by both GPT-5 and GPT-5.2, the GPT-5 variant wins on 24 (44\%) and solves all 3 problems that the GPT-5.2 MI agent could not.
$\mu$CUTLASS variants reach over 40\% of problems at $\ge 2\times$ within the first 5--10~attempts, while non-$\mu$CUTLASS variants plateau lower and converge more slowly.

The bottom row shows GPT-5.2, where the balance shifts.
The MI agent already has 63\% of problems improving over PyTorch, with 47\% at $\ge 2\times$.
$\mu$CUTLASS pushes coverage at $\ge 2\times$ from 47\% to 64\% and 37\% above $4\times$.
SOL guidance alone reaches 85\% improving over PyTorch, the highest solve rate of any single-component variant, but fewer reaching extreme speedups.
The combination reaches a median of $3.39\times$ with 25 problems above $4\times$, but does not exceed $\mu$CUTLASS alone---coverage at $\ge 2\times$ dips slightly (61\% vs.\ 64\%). The diminishing marginal return of SOL guidance atop $\mu$CUTLASS is 
notable.
We hypothesize that the strongest model already optimizes 
effectively with the DSL, and adding steering introduces overhead 
without new optimization insights.
All four GPT-5.2 variants converge rapidly, with $\mu$CUTLASS variants reaching their plateaus within the first few attempts.

\subsubsection{SOL Guidance Implementation: Orchestrated vs In-Prompt}
\label{sec:results:rq3}

\begin{figure}[!htbp]
\centering
\begin{minipage}[t]{0.495\linewidth}
\centering
\includegraphics[width=\linewidth]{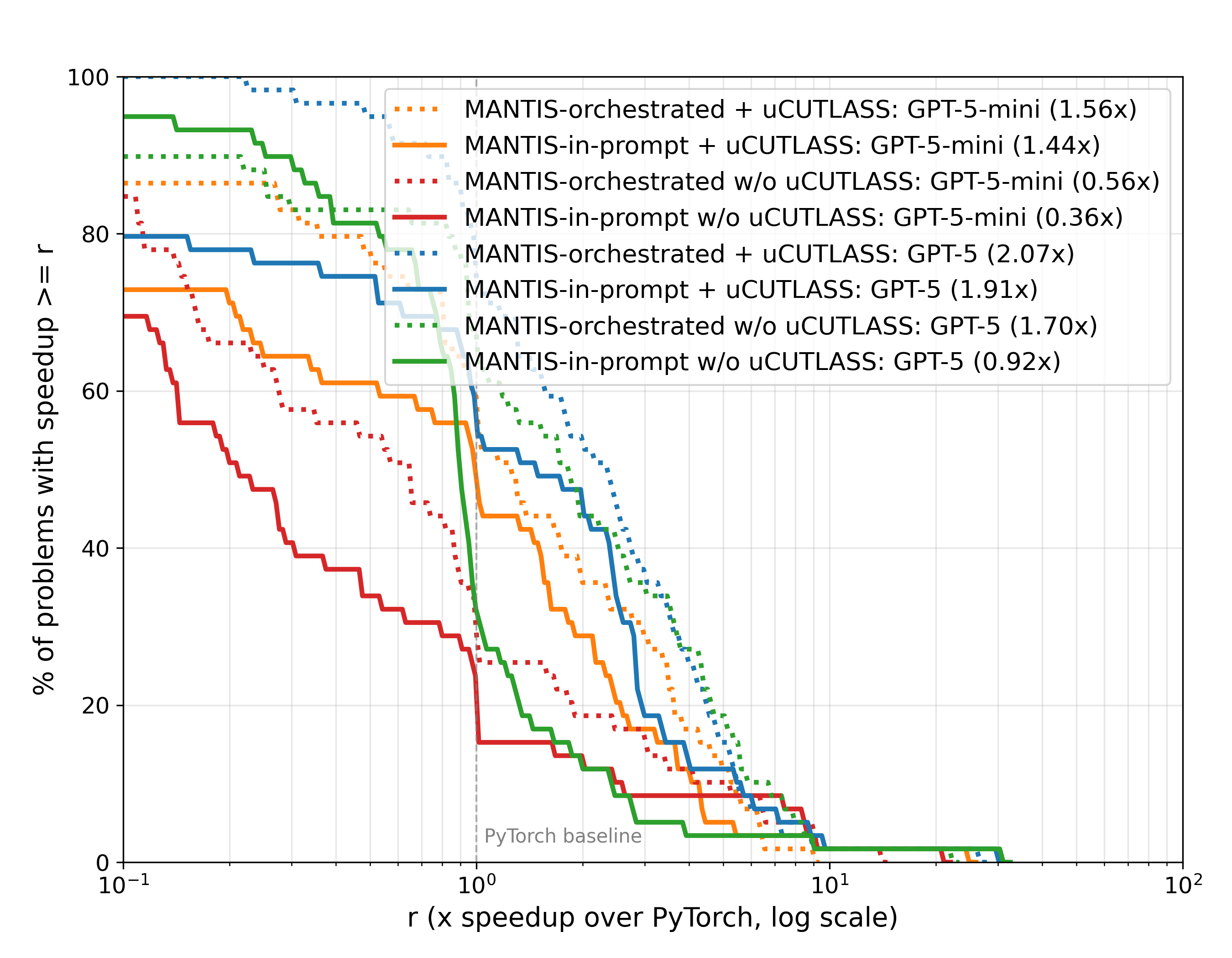}
\\[2pt]{\small (a) GPT-5-mini and GPT-5}
\end{minipage}\hfill
\begin{minipage}[t]{0.495\linewidth}
\centering
\includegraphics[width=\linewidth]{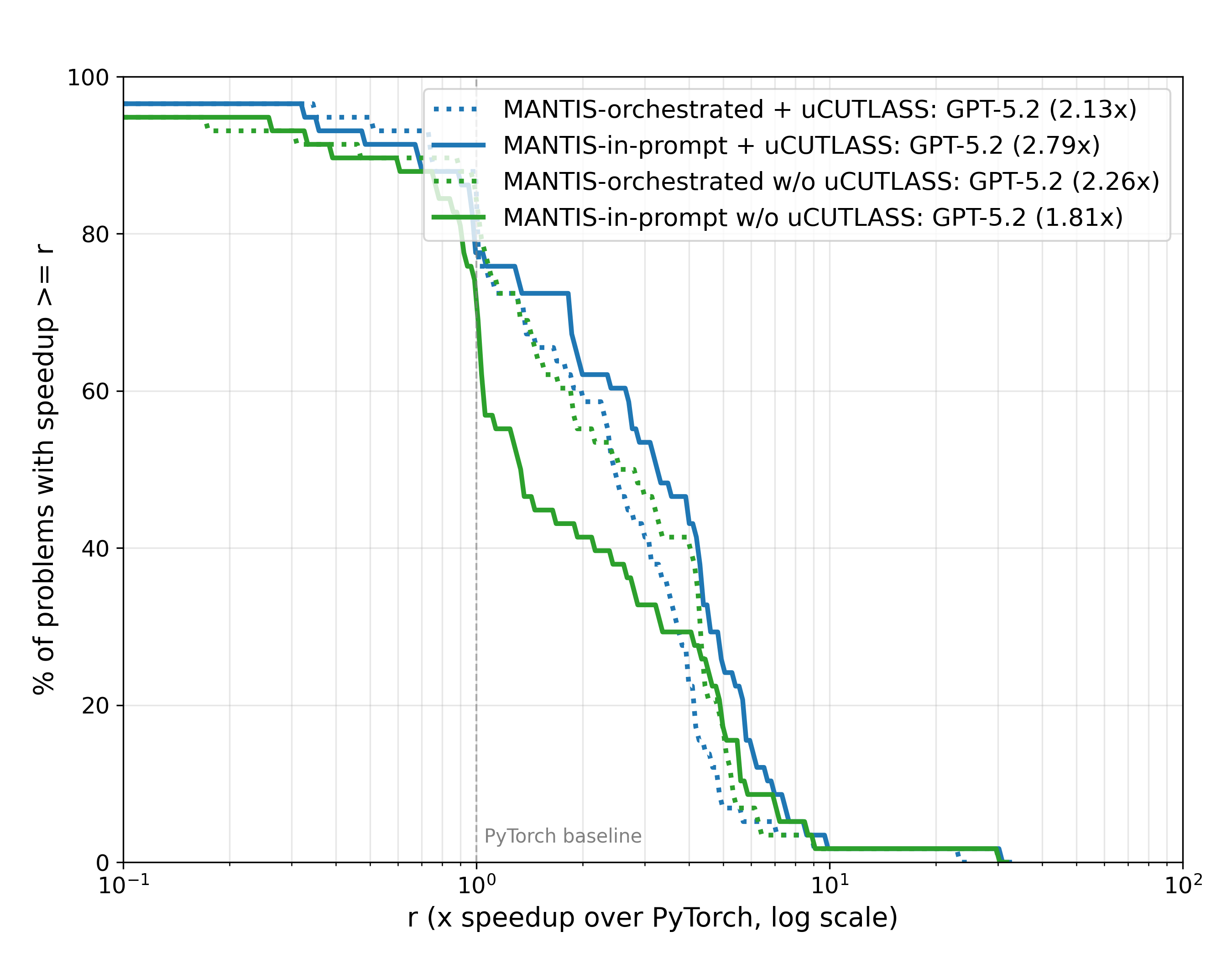}
\\[2pt]{\small (b) GPT-5.2}
\end{minipage}
\caption{Orchestrated vs.\ in-prompt SOL guidance implementation is shown here. Dotted = orchestrated, solid = in-prompt. Same color = same setting.}
\label{fig:orch_vs_prompt}
\end{figure}

As described in Section~\ref{sec:method:variants}, SOL-guided steering can be implemented as structured text in the agent's prompt (\emph{in-prompt}) or as a multi-phase orchestration pipeline (\emph{orchestrated}).
Figure~\ref{fig:orch_vs_prompt} compares these two forms across all three model tiers, with and without $\mu$CUTLASS.

Figure~\ref{fig:orch_vs_prompt}(a) shows Fast-$p$ curves for GPT-5-mini and GPT-5.
We use the signed-area metric (Section~\ref{sec:method:metrics}) to quantify the gap between two Fast-$p$ curves. A positive value means orchestrated's Fast-$p$ curve lies higher.
Without $\mu$CUTLASS, orchestrated variant dominates on both the models, but the gap is far larger on GPT-5 with signed area $+1.25$ vs. $+0.22$ for GPT-5-mini. We hypothesize that the stronger model can better exploit the structured multi-step pipeline, whereas GPT-5-mini's weaker reasoning limits the benefit of orchestration.
With $\mu$CUTLASS the gap narrows for both models. The gap for GPT-5 drops from $+1.25$ to $+0.59$, and GPT-5-mini stays near-flat ($+0.24$). The DSL absorbs much of the implementation burden that orchestration was compensating for, reducing the advantage of the multi-step pipeline regardless of model strength.

Figure~\ref{fig:orch_vs_prompt}(b) shows Fast-$p$ curves for GPT-5.2, where the pattern partially reverses.
Without $\mu$CUTLASS, orchestrated variant still remains higher ($+0.37$).
With $\mu$CUTLASS, however, in-prompt variant is ahead ($-0.87$). The curves cross around $r \approx 1.5$ and in-prompt's advantage grows at higher thresholds.
The reversal suggests that the rigid orchestration pipeline constrains a model whose inherent planning capabilities already exceed the imposed structure. With $\mu$CUTLASS handling implementation details, the strongest model benefits more from autonomy than from externally imposed steering.

\begin{description}
\item [RQ3: SOL Guidance Implementation.] The results indicate that the best implementation form is model-dependent. Orchestrated steering benefits weaker and mid-tier models by structuring the optimization process, while in-prompt steering is preferable for the strongest model when paired with $\mu$CUTLASS.
\end{description}

\subsubsection{Component Ablations}
\label{sec:results:ablations}

\begin{figure}[!htbp]
\centering
\begin{minipage}[t]{0.325\linewidth}
\centering
\includegraphics[width=\linewidth]{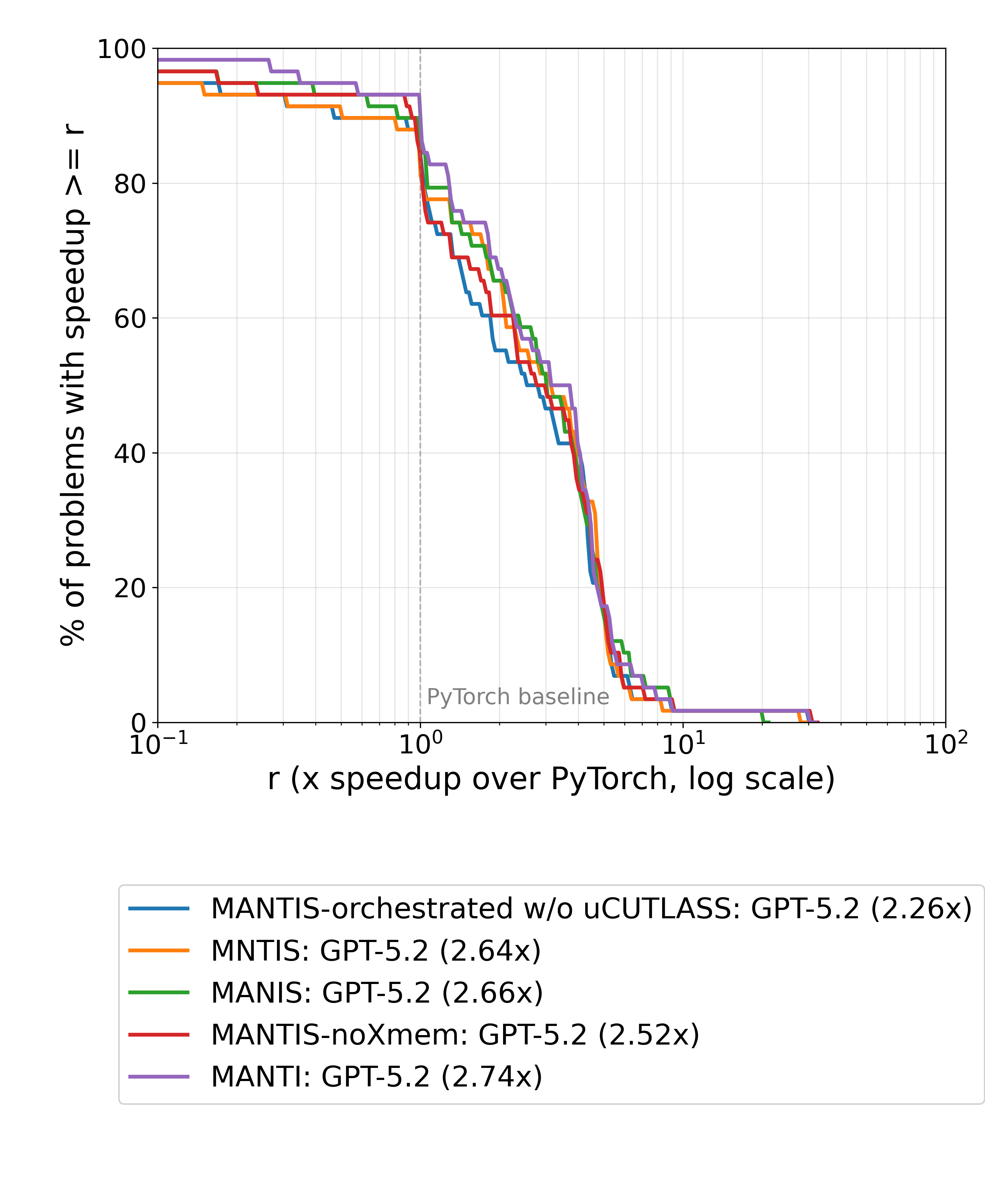}
\\[2pt]{\small (a) GPT-5.2, no $\mu$CUTLASS}
\end{minipage}\hfill
\begin{minipage}[t]{0.325\linewidth}
\centering
\includegraphics[width=\linewidth]{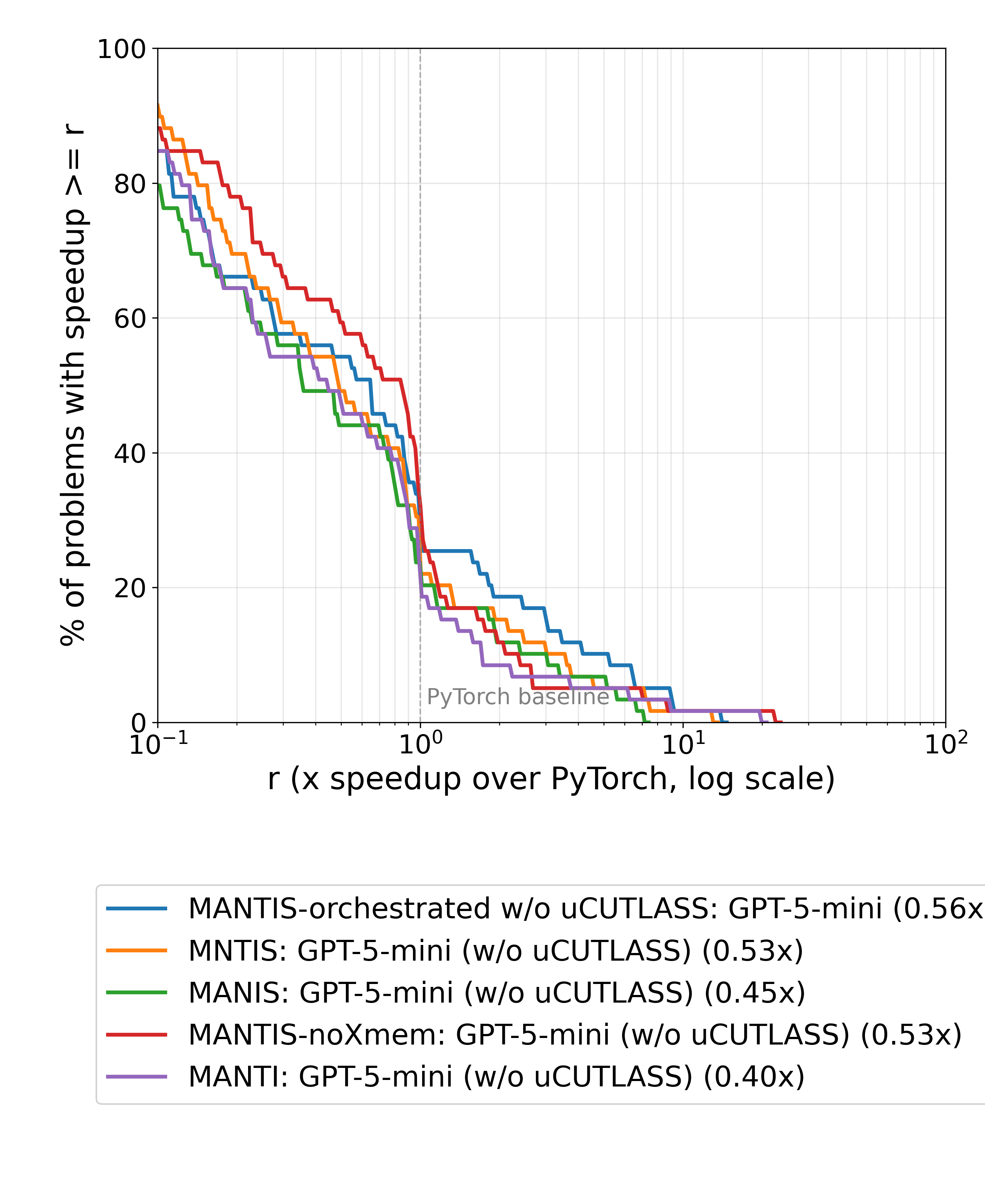}
\\[2pt]{\small (b) GPT-5-mini, no $\mu$CUTLASS}
\end{minipage}\hfill
\begin{minipage}[t]{0.325\linewidth}
\centering
\includegraphics[width=\linewidth]{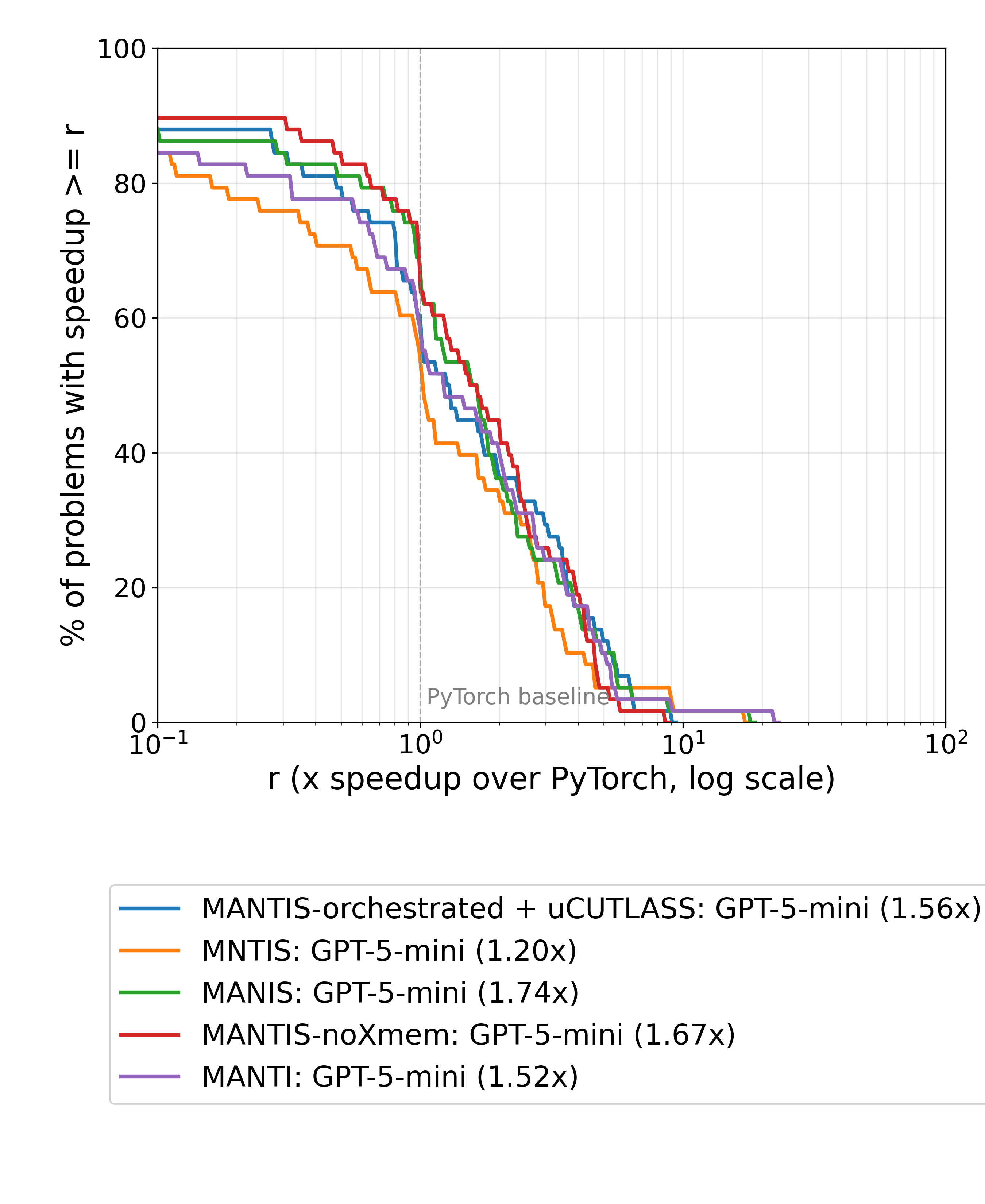}
\\[2pt]{\small (c) GPT-5-mini, with $\mu$CUTLASS}
\end{minipage}
\caption{Component ablation results are shown here. Fast-$p$ curves for full MANTIS and four single-component ablations on configurations where SOL guidance provides benefits.
}
\label{fig:ablations}

\end{figure}

We further study the importance of individual orchestration components by evaluating four single-component ablations (Table~\ref{tab:ablations}), each removing one of Analyze, Triage, Summarize, or cross-problem memory, while keeping the same attempt budget. We conducted experiments using GPT-5.2 and GPT-5-mini for this ablation study, considering GPU resource and cost constraints.
Since Section~\ref{sec:results:rq1rq2} showed that SOL guidance provides diminishing returns on GPT-5.2 when paired with $\mu$CUTLASS, we focus on the configurations where orchestration matters, namely GPT-5.2 without $\mu$CUTLASS and GPT-5-mini with and without $\mu$CUTLASS.
Figure~\ref{fig:ablations} shows Fast-$p$ curves and also the geomean speedups for each variant in the legend.

On GPT-5.2 without $\mu$CUTLASS (a), removing any single component is a wash. The strong model already plans well on its own, and the limitation to higher speedup lies elsewhere.
In contrast, on GPT-5-mini without $\mu$CUTLASS (b), every component matters. Removing any one causes a noticeable Fast-$p$ drop, with Triage and Summarize being the most damaging. The mini model benefits from the full orchestration structure. 
On GPT-5-mini with $\mu$CUTLASS (c), the picture is mixed. Removing Analyze (SOL analysis) hurts, but other ablations are a wash. The DSL absorbs enough of the implementation burden that most orchestration overhead no longer pays for itself.

These results suggest that multi-phase orchestration is not uniformly beneficial. Its value depends on the gap between what the model can do on its own and what the task demands. When that gap is large (weak model, no DSL), orchestration fills it and every component is helpful. As either model capability or tooling improves, the gap shrinks and orchestration components become less useful. As future work, this motivates a hybrid approach between in-prompt and orchestrated steering that adaptively selects MANTIS components based on model capability and available tooling.

\subsection{SOL-Guided Budget Scheduling}
\label{sec:results:rq4}

As discussed in Sections~\ref{sec:agent:sol_stopping} and~\ref{sec:agent:sol_scheduling}, SOL estimates enable budget scheduling policies that stop allocating attempts/iterations to a problem once its best kernel is close enough to the theoretical limit.
We evaluate two complementary stopping criteria via offline replay of existing run logs:
a \emph{SOL-headroom threshold}~$\epsilon$, which stops allocating resources to a problem once its best kernel beats PyTorch and satisfies $t_\mathrm{best} \le (1{+}\epsilon)\,t_\mathrm{SOL}$, and
a \emph{no-progress window}~$w$, which removes a problem whose best speedup has not improved for $w$ consecutive attempts while already ahead of PyTorch.
We explore each criterion independently and in combination (a problem is stopped when either criterion fires).
We use the tighter FP16 SOL bound for scheduling.
Some solutions may use FP32/TF32 arithmetic and therefore not come close to the FP16 ceiling.

\subsubsection{Independent Parameter Sweeps}

\begin{figure}[!htbp]
\centering
\includegraphics[width=\linewidth]{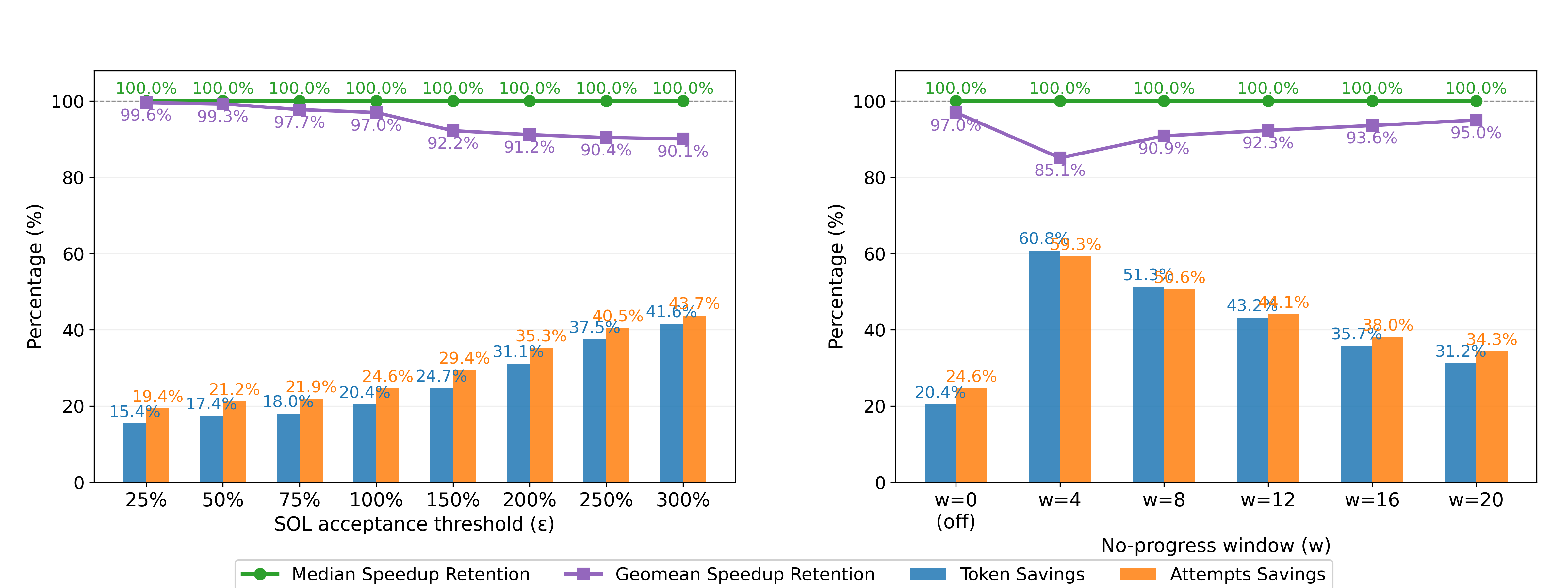}
\caption{Independent parameter sweeps for GPT-5.2 $\mu$CUTLASS + SOL-guided variant are shown here. (a)~SOL-headroom threshold $\epsilon$ with no-progress window off ($w{=}0$). (b)~No-progress window $w$ with SOL-headroom threshold fixed at $\epsilon{=}100\%$. Bars show token and attempts savings. Lines show median and geomean speedup retention.}
\label{fig:sched_indep}
\end{figure}

Figures~\ref{fig:sched_indep}~(a) and~(b) show the effects of sweeping both the parameters in isolation on GPT-5.2 with $\mu$CUTLASS + SOL-guided steering.
Figure (a) sweeps $\epsilon$ with no-progress stopping disabled ($w{=}0$). Savings begin at the lowest threshold: $\epsilon{=}25\%$ already saves 15\% of tokens with 99.6\% geomean retention. Savings grow steadily with $\epsilon$: 20\% at $\epsilon{=}100\%$ with geomean retention at 97\%, reaching 42\% at $\epsilon{=}300\%$ where geomean retention drops to 90\%. Median retention stays at 100\% throughout.
The savings come entirely from problems whose best kernel has reached near SOL.
Figure~(b) sweeps $w$ with the SOL-headroom threshold fixed at $\epsilon{=}100\%$.
Adding $w{=}4$ increases savings to 61\%, because the saturation detector catches many problems that SOL-headroom alone misses, though geomean retention drops to 85\%.
Larger windows ($w{=}12$--$20$) are more conservative as they give the agent more attempts to break through a plateau before stopping, trading off savings for higher retention (e.g., $w{=}16$ retains 94\% geomean with 36\% savings).

\subsubsection{Combined Pareto Analysis}

\begin{figure}[!htbp]
\centering
\includegraphics[width=0.98\linewidth]{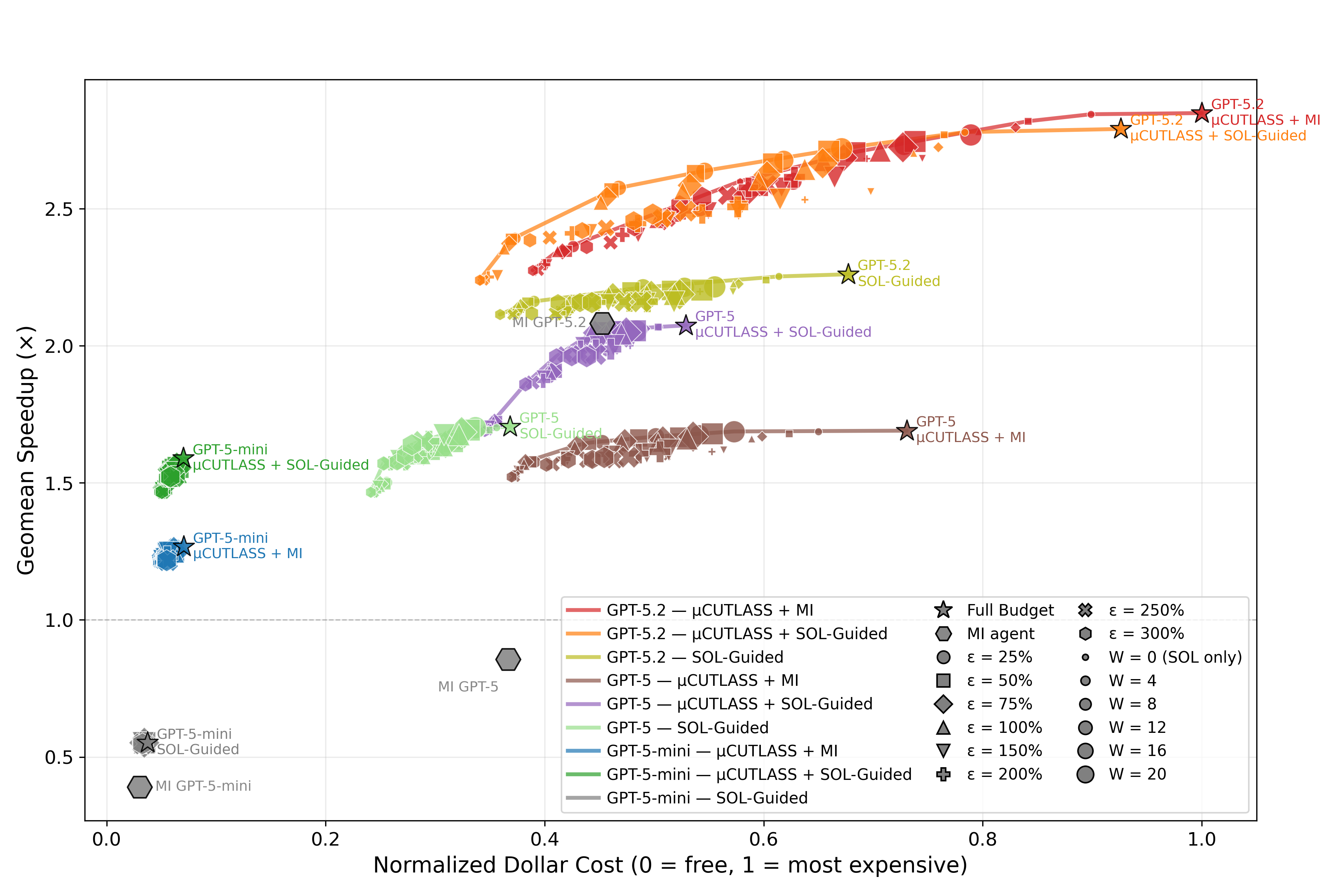}
\caption{Scheduler policy Pareto frontiers: normalized dollar cost vs.\ geomean speedup for nine variants across three per model tier are shown here. Each point is one $(\epsilon, w)$ combination. Marker shape and size encode $\epsilon$ and $w$, respectively. Star markers show the exhaustive baseline. Hexagons show MI agent reference points. Roofline envelopes trace the upper convex hull of each variant's frontier.}
\label{fig:pareto}
\end{figure}

To understand how the two parameters interact across variants, we sweep $\epsilon \in \{25\%, 50\%, \ldots, 300\%\}$ and $w \in \{0, 4, 8, 12, 16, 20\}$ and plot all $(\epsilon, w)$ combinations on a Pareto frontier of normalized dollar cost versus geomean speedup (Figure~\ref{fig:pareto}).
Dollar cost is estimated by multiplying total token count by the per-model input-token price (\$0.25/M for GPT-5-mini, \$1.25/M for GPT-5, \$1.75/M for GPT-5.2).
We normalize all points relative to the most expensive variant's fixed (no-stopping) run so that the $x$-axis ranges from 0 (free) to 1 (most expensive).
For each variant, we draw a roofline-style envelope connecting the outermost points to form a piecewise-linear upper bound on the cost versus speedup trade-off.
We make following four observations from the results in Figure~\ref{fig:pareto}.

\begin{itemize}
  \item \emph{Scheduling turns each variant into a cost vs.\ speedup frontier.}
    The scheduler does not produce a single operating point. It exposes a continuum of budget and speedup trade offs for each variant.
    For GPT-5.2 $\mu$CUTLASS + SOL-guided, the exhaustive point is at approximately $(0.93, 2.65\times)$ in Figure~\ref{fig:pareto}. A moderate policy at $\epsilon{=}25\%$ and $w{=}16$ moves to approximately $(0.62, 2.55\times)$, which corresponds to a $33\%$ token reduction with $96\%$ geomean speedup retention.

  \item \emph{$\mu$CUTLASS and SOL-guided steering lift the frontier within a model tier.}
    On GPT-5.2, the $\mu$CUTLASS + SOL-guided curve lies above the $\mu$CUTLASS + MI curve over much of the plotted range. At comparable normalized cost, around $0.62$ to $0.64$, $\mu$CUTLASS + SOL-guided reaches about $2.55\times$, while $\mu$CUTLASS + MI reaches about $2.39\times$.
    The same pattern appears on GPT-5. At normalized cost $\approx 0.53$, $\mu$CUTLASS + SOL-guided reaches $2.07\times$, whereas $\mu$CUTLASS + MI reaches $1.60\times$. This indicates that agent design determines the vertical position of the frontier within a model tier, while scheduling selects operating points along that frontier.

  \item \emph{The most useful scheduler lever depends on model tier.}
    On GPT-5-mini $\mu$CUTLASS + SOL-guided, sweeping $\epsilon$ alone from $25\%$ to $300\%$ changes the frontier only slightly, from approximately $(0.070, 1.48\times)$ to $(0.066, 1.46\times)$. In contrast, setting $w{=}4$ moves the frontier to approximately $(0.049, 1.40\times)$ and yields about $30\%$ token savings.
    On GPT-5 and GPT-5.2, both $\epsilon$ and $w$ move points substantially, which produces a broader range of achievable trade offs. We study this variant-dependent behavior further in Section~\ref{sec:results:rq4:gain}, where we evaluate different $(\epsilon, w)$ combinations for different variants.

  \item \emph{SOL guidance enables principled scheduling.}
    Each variant contributes $48$ $(\epsilon, w)$ policies, so the scheduler design space is already large even before model selection. SOL guidance provides a measurable ceiling that turns scheduling into a principled decision: stop when additional search is unlikely to close the remaining gap to the theoretical limit.
\end{itemize}

These frontiers summarize geomean behavior over 59 problems. Individual problems may favor different operating points, so an important next step is to understand which problem characteristics predict the best scheduling regime.

\subsubsection{Efficiency Gain}
\label{sec:results:rq4:gain}

We use the \emph{efficiency gain} metric defined in Section~\ref{sec:method:metrics} to quantify the trade-off between savings and speedup retention.
Any gain above $1\times$ indicates that the tokens saved outweigh the speedup lost.

\begin{figure}[!htbp]
\centering
\includegraphics[width=\linewidth]{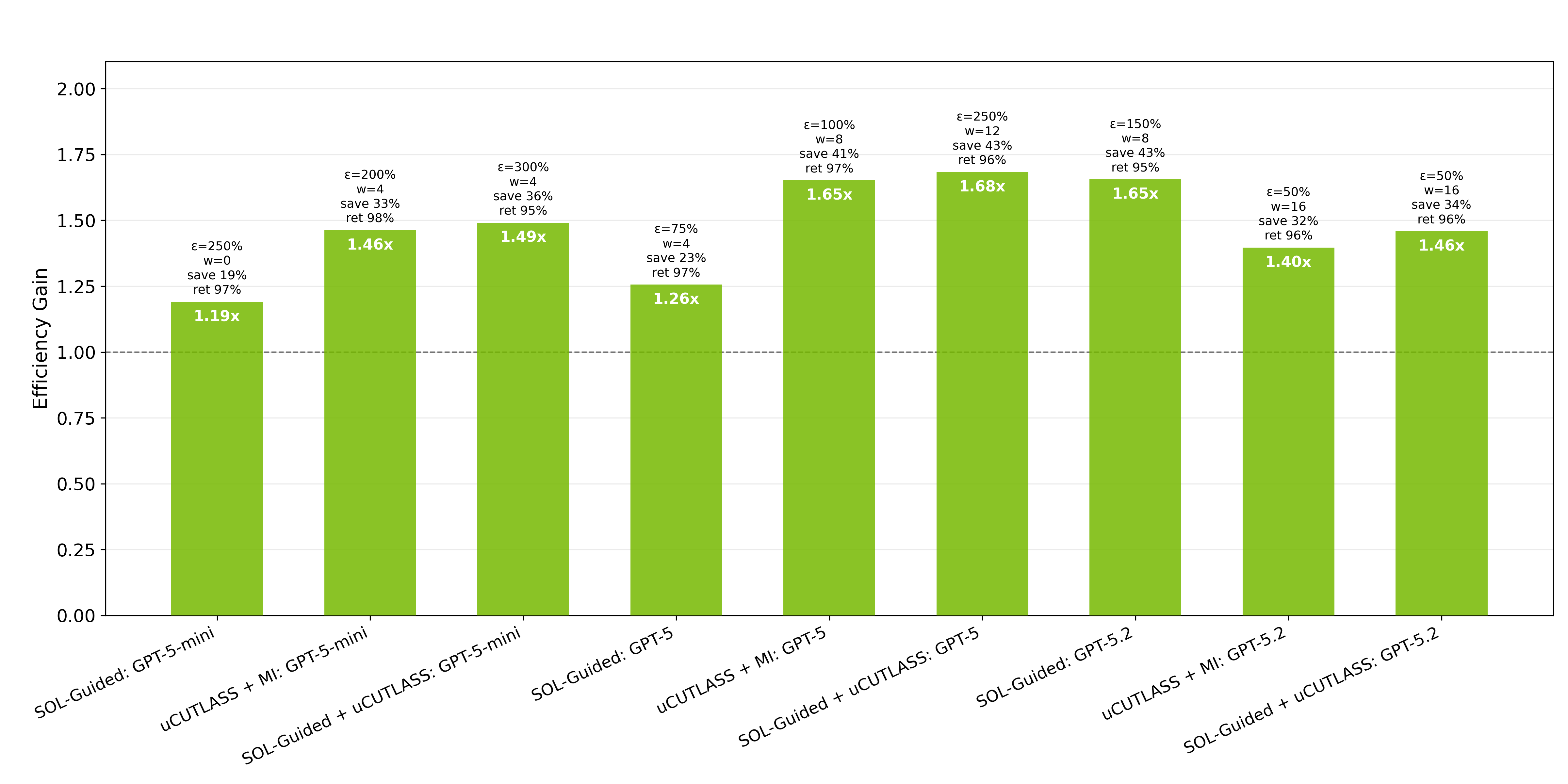}
\caption{Best scheduler policy per variant is shown here. For each variant, the $(\epsilon, w)$ combination maximizing efficiency gain is selected subject to $\ge 95\%$ geomean retention. Annotations show the chosen parameters and token savings.}
\label{fig:best_combo}
\end{figure}

Figure~\ref{fig:best_combo} selects the $(\epsilon, w)$ combination that maximizes efficiency gain for each variant, subject to a $\ge 95\%$ geomean retention constraint.
The strongest result is GPT-5 $\mu$CUTLASS + SOL at $1.68\times$ gain ($\epsilon{=}250\%$, $w{=}12$ with 43\% savings and 96\% speedup retention).
GPT-5 $\mu$CUTLASS + MI follows at $1.65\times$ ($\epsilon{=}100\%$, $w{=}8$ with 42\% savings).
On GPT-5.2, the $\mu$CUTLASS + SOL variant achieves $1.46\times$ gain ($\epsilon{=}50\%$, $w{=}16$ with 34\% savings), while GPT-5.2 SOL-Guided reaches $1.65\times$ ($\epsilon{=}150\%$, $w{=}8$ with 43\% savings).
GPT-5-mini shows smaller but consistent gains with $1.49\times$ for $\mu$CUTLASS + SOL ($\epsilon{=}300\%$, $w{=}4$ with 36\% savings) and $1.46\times$ for $\mu$CUTLASS + MI ($\epsilon{=}200\%$, $w{=}4$ with 33\% savings).

The selected policies show a wider spread than before with the corrected FP16 SOL, with $\epsilon$ ranging from $50\%$ to $300\%$ across variants. Weaker models prefer higher $\epsilon$ thresholds since their kernels rarely reach near the FP16 SOL, while stronger models benefit from tighter thresholds. The preferred window still varies by variant: GPT-5 prefers $w{=}8$--$12$, GPT-5-mini prefers $w{=}4$, and GPT-5.2 variants prefer $w{=}16$. A useful direction for future work is to predict a good scheduler policy before running experiments.

Based on these results and analysis, we answer RQ4 from Section~\ref{sec:evaluation}.

\begin{description}
\item[RQ4: SOL-guided budget scheduling.]
SOL-guided budget scheduling converts each fixed-budget agent into a cost vs.\ speedup frontier and often removes wasted late-stage search near SOL. Across all nine variant--model configurations, the best policy under the $\ge 95\%$ geomean retention constraint saves 19--43\% of tokens.
On GPT-5.2 $\mu$CUTLASS + SOL-guided, a moderate policy with $\epsilon{=}25\%$ and $w{=}16$ moves from approximately $(0.93, 2.65\times)$ to $(0.62, 2.55\times)$, which corresponds to 33\% token savings with 96\% speedup retention.

\end{description}

\subsection{SOL-Guided Integrity Checking}
\label{sec:results:rq5}

As discussed in Sections~\ref{sec:agent:sanity_checking} and~\ref{sec:agent:sol_integrity}, correctness tests and agent instructions do not fully specify what counts as a legitimate kernel, and the space of possible gaming strategies is open-ended. We therefore perform an offline post-analysis using SOL-guided integrity checking, rather than trying to enumerate all exploit types in advance.
Our three-step pipeline uses SOL in its first two stages. An \emph{SOL-ceiling detector} flags runtimes faster than the FP16 SOL bound, and an \emph{LLM-based game detector} (LGD) reviews the code together with the SOL report and labels attempts as \emph{No Issues}, \emph{Minor Issues}, or \emph{Gaming}, which we further divide into Original and Inherited Gaming. A third static \emph{PyTorch-only detector} catches attempts that only compose library calls such as \texttt{torch.mm} or \texttt{F.conv2d} without implementing custom CUDA, CUTLASS, or $\mu$CUTLASS kernels.
Because this review is performed offline, the agent cannot revise attempts that are nearly correct but have minor issues. We therefore accept attempts labeled \emph{No Issues} or \emph{Minor Issues}, since their performance appears to come from the intended optimization rather than from gaming, and exclude the rest.

\begin{figure}[!htbp]
  \centering
  \includegraphics[width=\linewidth]{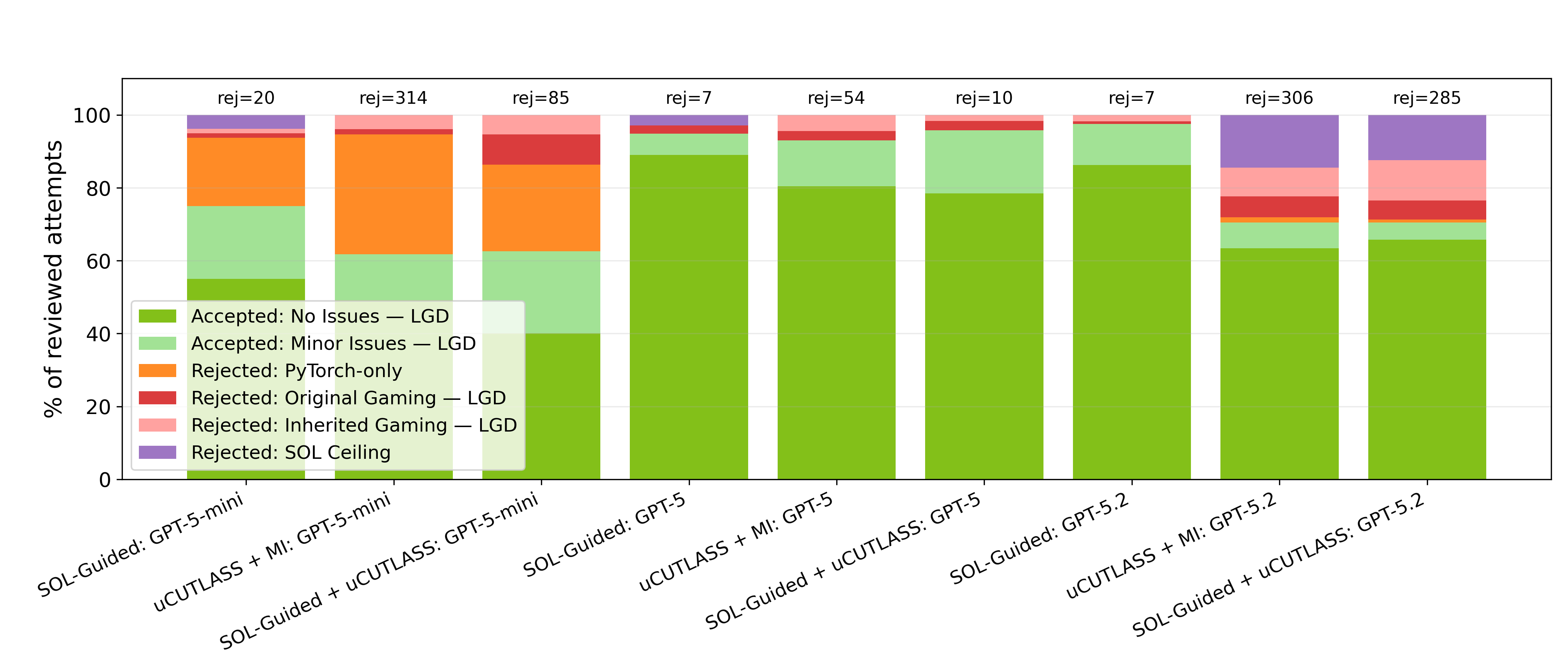}
  \caption{Review outcome composition for each variant. Six-band stacked bars show No Issues (accepted), Minor Issues (accepted), SOL Ceiling (rejected), PyTorch-only (rejected), Original Gaming (rejected), and Inherited Gaming (rejected).}
  \label{fig:integrity_outcomes}
  \end{figure}

\textbf{Outcome distributions.}
Figure~\ref{fig:integrity_outcomes} shows the outcome distribution for each agent variant.
The SOL-ceiling detector (purple) fires less frequently, confirming that most gaming attempts are not fast enough to beat the physical bound. It is most visible on SOL-guided GPT-5-mini, where a few attempts produce implausibly fast runtimes.
The PyTorch-only detector (orange) is particularly active on $\mu$CUTLASS + MI variants, where the agent achieves high speedups through PyTorch operator fusion without writing custom kernels.
We suspect that less capable models fall back to PyTorch library calls after failing to produce correct custom code.
These solutions are valid but do not represent the custom kernel development, which is the focus of this evaluation.
More capable models exhibit higher gaming rates detected by LGD.
On GPT-5.2, the $\mu$CUTLASS + MI variant triggers 306 exclusions, compared to 113 for MI on GPT-5-mini.
The inherited-gaming band (pink) is notably larger on stronger-model variants, revealing that once a model discovers a gaming shortcut in the optimization process, it does not revert to legitimate strategies.

From Section~\ref{sec:results:rq3} we know that orchestrated variant is used for SOL-guided agents except for GPT-5.2. Based on this information, we observe that SOL-guided orchestrated variants show markedly less gaming across all tiers (3--35 exclusions), suggesting that structured steering discourages shortcut discovery.

\begin{figure}[!htbp]
  \centering
  \includegraphics[width=\linewidth]{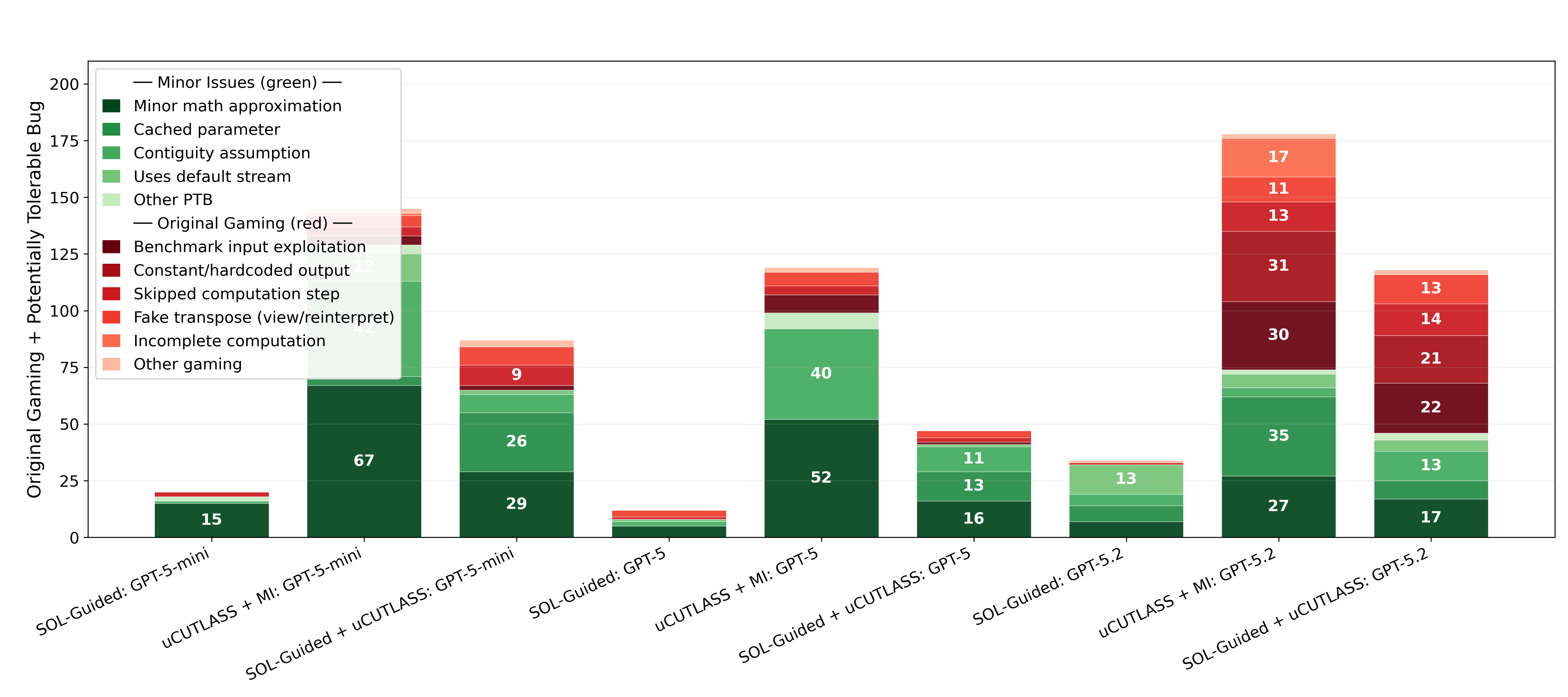}
  \caption{LGD category breakdown for each variant is shown here. Original Gaming  and Minor Issues subcategories are stacked.}
  \label{fig:integrity_categories}
\end{figure}

\textbf{LGD category breakdown.}
To understand what the LGD contributes within the broader integrity pipeline, Figure~\ref{fig:integrity_categories} breaks down all categories assigned by the LGD, the only non-static detector in our pipeline.
LLM classifies an attempt as \emph{Minor Issue} rather than \emph{Gaming} when it substantially preserves the intended computation but contains a small correctness or generality bug whose effect on performance is unlikely to be material.
Among Minor Issues (green shades):
(1) \emph{minor math approximation} where the kernel has a subtle math or precision difference (e.g., slightly different normalization order, missing epsilon term) that still passes the correctness check within tolerance. The kernel performs the intended computation with a minor flaw, unlike gaming attempts that skip computation entirely.
(2) \emph{cached parameter} where the kernel caches weights or biases keyed on shape or pointer rather than content, so values stay correct during benchmarking but would go stale if weights were updated in-place.
(3) \emph{contiguity assumption} where the kernel assumes contiguous memory layout. It computes correctly on the contiguous tensors provided by the test harness but would fail on sliced or transposed views with non-standard strides.
(4) \emph{uses default stream} where the kernel uses the default CUDA stream instead of PyTorch's current stream, introducing a latent data race that does not manifest during single-stream benchmarking.

Among original-gaming flags (red shades):
(1) \emph{benchmark input exploitation} where the kernel replaces a complex function with a linear or constant fit calibrated to the specific benchmark input shape,
(2) \emph{constant/hardcoded output} where the kernel ignores its input entirely and returns a pre-computed or cached tensor,
(3) \emph{skipped computation step} where the kernel omits a required pipeline stage such as dropout, bias addition, or activation clamping, producing a faster but incorrect result,
(4) \emph{fake transpose (view/reinterpret)} where the kernel uses \texttt{view} or \texttt{as\_strided} to reinterpret memory layout instead of performing a real data transpose (this can be acceptable if the downstream application handles non-contiguous layouts, but we conservatively mark it as gaming), and
(5) \emph{incomplete computation} where the kernel computes only a prefix or sub-sample and fills the remainder with zeros.

Some gaming types are model-specific while others are generic.
\emph{Constant/hardcoded output} concentrates on GPT-5.2 (104--139 flags per variant) and is rare on weaker models. Constructing a cached result that passes the correctness check requires sophisticated reasoning that weaker models do not exhibit.
\emph{Fake transpose} is associated with $\mu$CUTLASS. On GPT-5.2, $\mu$CUTLASS + SOL produces 44 fake-transpose flags compared to 0 for the SOL-only variant, as the DSL facilitates view-based layout manipulation.
\emph{Skipped computation step} is observed across all models and variants, indicating a generic gaming risk.

\begin{figure}[!htbp]
  \centering
  \includegraphics[width=\linewidth]{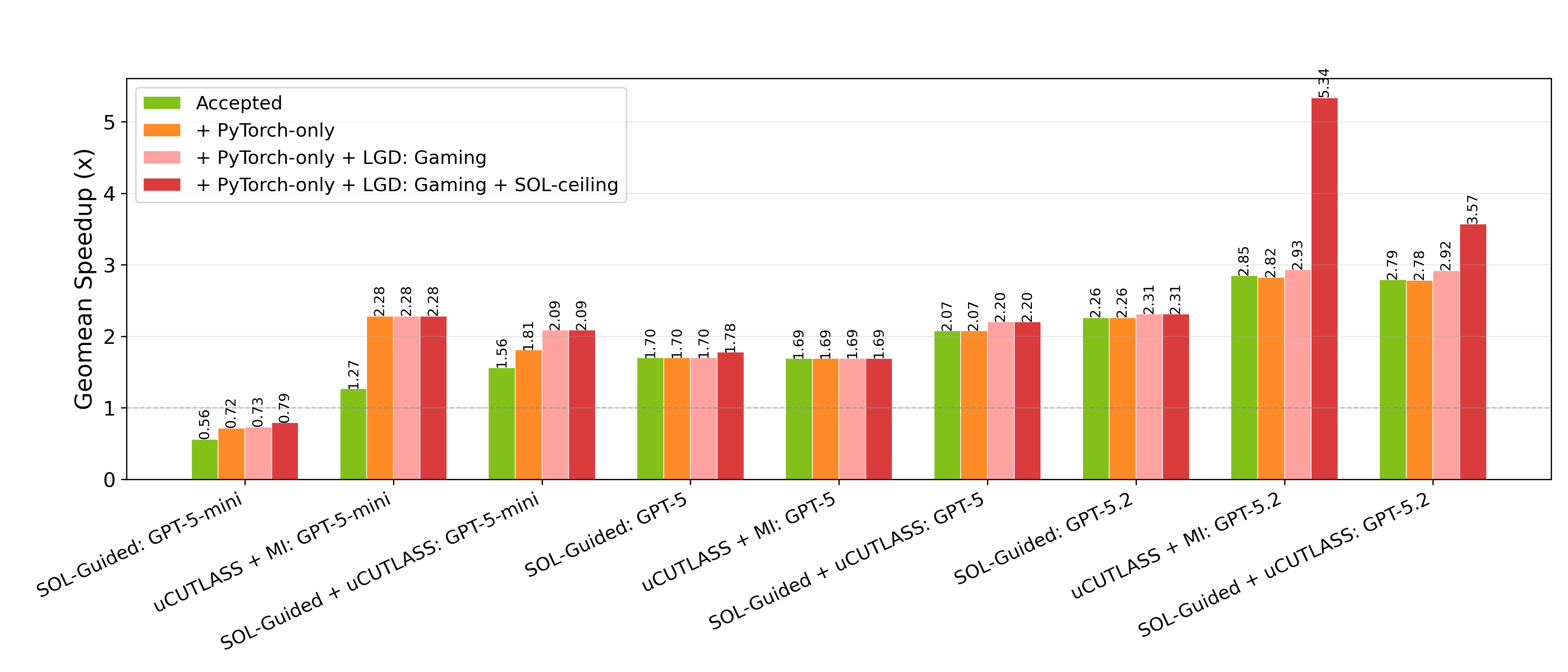}
  \caption{speedup inflation without the integrity pipeline. Green bars show the filtered results. Orange, light red, and red bars additionally allow PyTorch-only, LGD-labeled gaming solutions, and SOL-ceiling detected solutions, respectively.}
  \label{fig:integrity_policy}
  \end{figure}

\textbf{Speedup inflation without integrity checking.}
Figure~\ref{fig:integrity_policy} compares the filtered geomean speedup against the speedups that would be reported if certain categories were not filtered out. Green bars show the filtered results, orange bars allow PyTorch-only solutions, light red bars allow LGD labeled gaming solutions, and red bars are fully unfiltered.
The gap between green and red quantifies the total speedup inflation that gaming and PyTorch-only solutions introduce. 

Without the PyTorch-only filter, reported speedups inflate substantially on $\mu$CUTLASS variants. For example, GPT-5-mini $\mu$CUTLASS + MI rises from $1.27\times$ to $2.28\times$, an 80\% inflation, because many of the fastest solutions are PyTorch operator fusion rather than custom kernels.
PyTorch-only flags also concentrate on more complex problems. All studied Level-3 problems have at least one PyTorch-only attempt (64 total), compared with 14 of 19 Level-2 problems (121 attempts) and 12 of 32 Level-1 problems (35 attempts). This suggests that when custom multi-operator fusion is difficult, the model often falls back to library-call composition. Although these solutions are functionally valid, we exclude them because our evaluation targets custom kernel development rather than library composition.

Without the game detector, the inflation is even larger on stronger models. GPT-5.2 $\mu$CUTLASS + MI rises from $2.85\times$ to $5.34\times$, a $1.9\times$ inflation driven largely by gaming exploits. SOL-guided variants are less affected but not immune. GPT-5.2 SOL-guided + $\mu$CUTLASS still rises from $2.79\times$ to $3.57\times$. Reporting unfiltered results would therefore substantially overstate the agent's kernel-writing ability.

\textbf{Prompt-level integrity guardrails.}
We tested whether prompt-level instructions could substitute for the integrity pipeline by repeating the GPT-5-mini experiments with explicit anti-PyTorch-only and anti-gaming clauses.
Table~\ref{tab:prompt_guardrails} shows the results.
The anti-PyTorch-only instruction significantly reduced PyTorch-only flags (e.g., from 345 to 51 for $\mu$CUTLASS + MI) but did not eliminate them.
Gaming flags were not consistently reduced. On $\mu$CUTLASS + MI, gaming actually increased from 50 to 95 attempts.
These findings reinforce that prompt-level guardrails alone are insufficient and a dedicated detection mechanism remains necessary.

\begin{table}[!htbp]
  \centering
  \small
  \caption{Prompt-level integrity guardrails on GPT-5-mini. Run~1 used the original prompt. Run~2 added anti-PyTorch-only and anti-gaming instructions.}
  \label{tab:prompt_guardrails}
  \begin{tabular}{lrrrr}
  \toprule
  & \multicolumn{2}{c}{PyTorch-only} & \multicolumn{2}{c}{Gaming (LGD)} \\
  \cmidrule(lr){2-3} \cmidrule(lr){4-5}
  Variant & Run~1 & Run~2 & Run~1 & Run~2 \\
  \midrule
  MI                          & 113 &  8 &  9 &  1 \\
  $\mu$CUTLASS + MI           & 345 & 51 & 50 & 95 \\
  SOL-Guided                  &  15 & 10 &  3 & 11 \\
  $\mu$CUTLASS + SOL-Guided   &  63 & 26 & 35 & 19 \\
  \bottomrule
  \end{tabular}
\end{table}

Based on these results and analysis, we next answer the last RQ from Section~\ref{sec:evaluation}.

\begin{description}
\item[RQ5: SOL-guided integrity checking.]
Integrity checking is essential for reliable evaluation because gaming strategies are open-ended and hard to enumerate in advance. SOL guidance helps by providing a first-principles signal of the work a legitimate kernel should perform. The SOL-ceiling detector alone fires infrequently, but the full pipeline removes 7--314 gaming or PyTorch-only attempts per variant and prevents reported geomean speedups from being inflated by up to $1.9\times$. Filtering is most important for $\mu$CUTLASS + MI and substantially less so for SOL-guided variants, suggesting that SOL-guided steering improves both optimization quality and integrity.
\end{description}


\subsection{Performance Stability Across Runs}
\label{sec:results:variation}
Multi-attempt LLM-based agent runs can produce different results when repeated, as calls to frontier models are non-deterministic.
This raises two natural questions. (1)~Are the improvements over baselines robust and relative rankings preserved, or did we get lucky with a favorable run? (2)~How does variability relate to model capability?

\begin{figure}[!htbp]
  \centering
  \includegraphics[width=\linewidth]{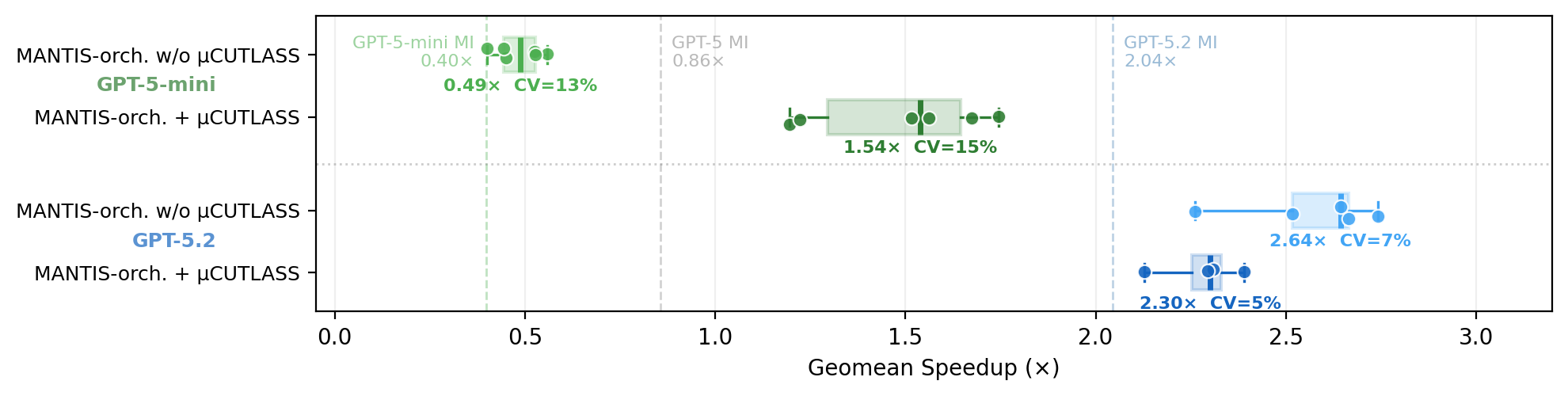}
  \caption{Run-to-run variation across model tiers. CV is the coefficient of variation ($\sigma/\mu$). GPT-5.2 groups contain $N{=}5$ (w/o $\mu$CUTLASS) and $N{=}4$ (+ $\mu$CUTLASS, fewer ablations were run) ablation variants. GPT-5-mini groups contain $N{=}6$ each: 5~ablation variants plus one independent repeat run.}
  \label{fig:variation}
\end{figure}

To gather evidence, we draw data from two sources.
First, the component ablations from Section~\ref{sec:results:ablations} produce nearby variants of the SOL-guided orchestrated agent. Together they provide an envelope of performance variation under prompt and component changes.
Second, we conducted an independent repeat of the GPT-5-mini experiments that used a different prompt with anti-gaming and anti-PyTorch-only instructions (as discussed in Section~\ref{sec:results:rq5}).
Figure~\ref{fig:variation} summarizes the observed spread across configurations spanning two model tiers and two agent variants.
Each dot is the geomean speedup of one configuration. Boxes show the interquartile range with whiskers extending to min/max.
Vertical dashed lines mark the MI baselines.

\textbf{(1) Performance gains persist despite run-to-run variation.}
Every GPT-5-mini configuration with $\mu$CUTLASS---including the lowest ablation variant (MNTIS at $1.20\times$) and the independent repeat run ($1.22\times$)---exceeds the GPT-5 MI baseline of $0.86\times$ by at least $40\%$.
The cross-run gap for $\mu$CUTLASS + SOL-guided ($1.56\times \to 1.22\times$, a $0.34\times$ shift) falls within the ablation envelope ($1.20$--$1.74\times$, a $0.55\times$ spread), indicating that the repeat run lies within the observed spread across nearby configurations.
Relative variant rankings are preserved across both the repeat run and the ablations: $\mu$CUTLASS + SOL-guided remains the strongest, and $\mu$CUTLASS consistently provides a large improvement over the corresponding non-DSL variant.
The main conclusions from the primary results therefore remain unchanged.

\textbf{(2) Variation decreases with model capability.}
GPT-5.2 variants cluster tightly: $5\%$ CV with $\mu$CUTLASS and $7\%$ without, across four and five configurations respectively.
The corresponding Fast-$p$ curves in Figure~\ref{fig:ablations}(a) are nearly indistinguishable.
GPT-5-mini shows wider variation ($13$--$15\%$ CV), consistent with weaker models exhibiting more sensitivity to agent/prompt configuration and generation randomness.

\subsection{Comparison to Prior Work}

\label{sec:results:sakana}
\begin{figure}[!htbp]
\centering
\includegraphics[width=0.75\linewidth]{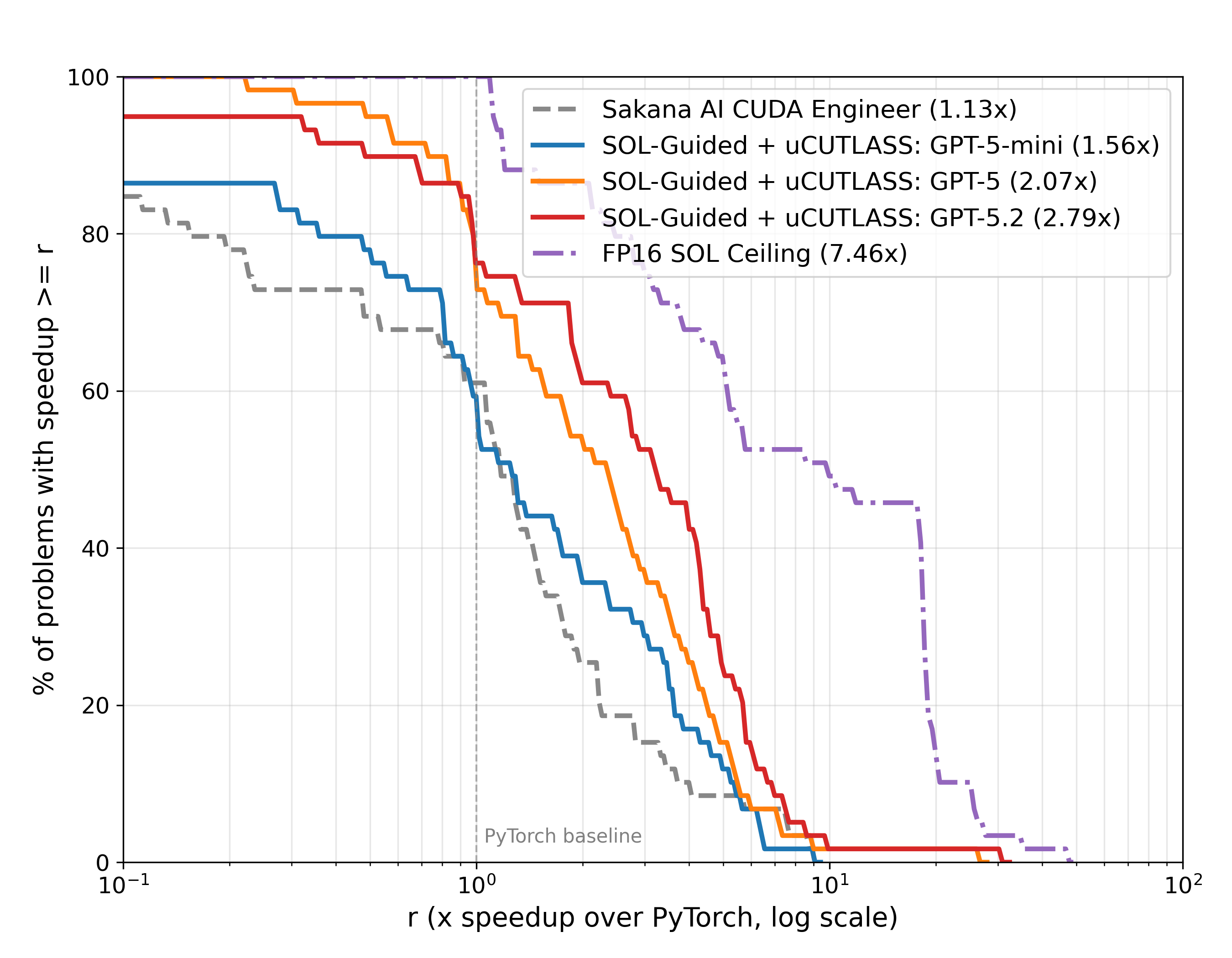}
\caption{Fast-$p$ curves comparing our $\mu$CUTLASS + SOL-guided variants across three model tiers against Sakana AI CUDA Engineer (dashed gray). The FP16 SOL curve (dashed purple) shows the theoretical speedup limit.}
\label{fig:sakana_fastp}
\end{figure}

Figure~\ref{fig:sakana_fastp} compares $\mu$CUTLASS + SOL-guided across three model tiers against the Sakana AI CUDA Engineer~\cite{lange2025aicudaengineer}, using the integrity-filtered kernels provided on huggingface as described in Section~\ref{sec:method:external_baseline}.
Because the software stack and runtime measurement methodology differ between the two evaluations, we compare against Sakana AI's reported speedups directly.

Of the 59~evaluation problems, 2 have no correct kernel in the Sakana archive.
Of the remaining 57, our LLM-based review accepts 52 kernels (28~No Issue, 24~Minor Issue) and rejects 5 (3~PyTorch-only, 2~Gaming), after evaluating 450 candidate kernels across the fallback review process.
The accepted Sakana kernels achieve a geomean speedup of $1.13\times$ over their own PyTorch baseline.
The 7~problems with no accepted kernel (2~missing, 5~rejected) are assigned speedup zero, counting against Sakana in the Fast-$p$ comparison.

The Fast-$p$ curves show a clear separation.
All three $\mu$CUTLASS + SOL-guided results provide higher speedups compared to the Sakana baseline.
Even GPT-5-mini with $\mu$CUTLASS + SOL-guided ($1.56\times$ geomean) substantially outperforms Sakana ($1.13\times$), with a higher fraction of problems reaching $\ge 2\times$ speedup.
GPT-5 ($2.07\times$) and GPT-5.2 ($2.79\times$) widen the gap further.
The comparison highlights that the SOL-guided steering and $\mu$CUTLASS DSL provide large gains beyond what a different frontier model alone achieves with an evolutionary search strategy.

If we select the fastest integrity-filtered kernel found for each problem across all our variants and runs, the resulting geomean speedup is $3.91\times$ ($4.53\times$ median), with all 59~problems solved. Figure~\ref{fig:sakana_fastp} also includes the \emph{FP16 SOL} curve (dashed purple), which reaches a $7.46\times$ geomean speedup. Closing this gap through adaptive variant selection, multi-variant ensembles, SOL-informed generation, or improved search strategies could yield substantial gains while keeping generation efficient.

\section{Conclusion}
\label{sec:conclusion}

We evaluated two methods to improve the efficiency of LLM-driven GPU kernel optimization across three model tiers (GPT-5-mini, GPT-5, GPT-5.2).
The first is raising the level of abstraction using a domain-specific language, $\mu$CUTLASS, that lets the model reason in an optimization space offering high-impact levers while hiding low-level implementation complexity. The grammar is compact enough to be learned entirely in context. The DSL alone turns GPT-5-mini's flat baseline from a $0.40\times$ regression to a $1.27\times$ geomean speedup, and benefits all three model tiers, with the largest gains on the weakest model.

The second is Speed-of-Light (SOL) guided optimization steering, which uses first-principles performance bounds to structure the agent's search. Combined with $\mu$CUTLASS, SOL-guided steering reaches $1.56\times$, $2.07\times$, and $2.79\times$ geomean speedup on GPT-5-mini, GPT-5, and GPT-5.2, respectively, enabling each model to match or exceed the next tier's raw-code baseline at lower inference cost.
We further employ SOL guidance to improve efficiency by allocating budget to problems that have higher optimization headroom. SOL-guided scheduling saves 19 to 43\% of tokens while retaining $\ge 95\%$ of geomean speedup, with the best configuration achieving a $1.68\times$ efficiency gain.

We found that game detection is crucial for LLM-based code optimization, as LLMs often find loopholes in specifications and produce kernels that appear fast without performing the intended computation. SOL guidance helps here as well, first as a static runtime ceiling to flag physically implausible results, and then to augment the specification for an LLM-based game detector. Without such detection, our results show that reported speedups can be inflated by up to $1.9\times$.

Future work includes automatically discovering compact DSL abstractions for new optimization domains, using per-problem characteristics to choose the most effective agent strategy and budget allocation, integrating integrity feedback into the optimization loop so that agents can correct gaming behavior during search, and applying first-principles performance bounds to steer LLM-based optimization in domains beyond GPU kernels.

\begin{ack}
We thank Hanfeng Chen, Vinod Grover, Sahil Modi, and Mark Stephenson for insightful discussions.
GenAI tools were used to create content in this paper.
\end{ack}

\bibliographystyle{ACM-Reference-Format}
\bibliography{references}

\appendix
\section{Appendix}
\label{sec:appendix}

\subsection{Full EBNF Grammar for $\mu$CUTLASS}
\label{sec:ucutlass:ebnf}

We include the full EBNF grammar used as the reference specification here.

\lstdefinestyle{ebnf}{
  basicstyle=\ttfamily\scriptsize,
  commentstyle=\color{rulecolor}\itshape,
  stringstyle=\color{processcolor},
  breaklines=true,
  columns=fullflexible,
  keepspaces=true,
  showstringspaces=false,
  numbers=left,
  numberstyle=\tiny\color{rulecolor},
  inputencoding=utf8,
  literate={µ}{{$\mu$}}1 {→}{{$\rightarrow$}}1 {↔}{{$\leftrightarrow$}}1 {×}{{$\times$}}1,
}

\begin{lstlisting}[style=ebnf,inputencoding=utf8,caption={$\mu$CUTLASS DSL grammar (EBNF).},label={lst:ucutlass-ebnf}]
(* µCUTLASS DSL Grammar (EBNF) *)
(* Clean, unquoted syntax - no string quotes needed except for custom expressions *)

(* TOP-LEVEL *)
start = kernel | pipeline ;
kernel = operation , { configuration } , { epilogue } ;

(* PIPELINES *)
pipeline = "pipeline(" , stage , { "," , stage } , ")" ;
stage = transform_stage | kernel_stage ;
kernel_stage = operation , { configuration } , { epilogue } ;
transform_stage = transpose_op ;

(* Transpose with optional dtype conversion:
 *   transpose(input, NCL, NLC)                - same dtype
 *   transpose(input, NCL, NLC, fp32, fp16)    - fp32 → fp16 conversion
 *   transpose(output, NLC, NCL, fp16, fp32)   - fp16 → fp32 conversion
 *   transpose(input, NCHW, NHWC)              - 4D layout swap (implemented by flattening HW)
 * Dtype conversion is fused with transpose (essentially free)
 *)
transpose_op = "transpose(" , TENSOR_REF , "," , LAYOUT_3D , "," , LAYOUT_3D , 
               [ "," , DTYPE , "," , DTYPE ] , ")" ;
TENSOR_REF = "input" | "output" | "weight" ;
(* NOTE: LAYOUT_3D also includes 4D logical layouts (NCHW/NHWC). *)
LAYOUT_3D = "NCL" | "NLC" | "NCHW" | "NHWC" ;

(* OPERATIONS *)
operation = gemm_op | batched_gemm_op | grouped_gemm_op 
          | conv2d_fprop | conv2d_dgrad | conv2d_wgrad
          | conv1d_fprop | depthwise_conv1d | group_conv1d
          | conv3d_fprop | conv3d_dgrad | conv3d_wgrad
          | depthwise_conv2d | group_conv2d | group_conv3d ;

gemm_op = "gemm()" ;
batched_gemm_op = "batched_gemm()" ;
grouped_gemm_op = "grouped_gemm(" , "expert_count=" , INTEGER , ")" ;

conv2d_fprop = "conv2d_fprop(" , "kernel_h=" , INTEGER , "," , "kernel_w=" , INTEGER , ")" ;
conv2d_dgrad = "conv2d_dgrad(" , "kernel_h=" , INTEGER , "," , "kernel_w=" , INTEGER , ")" ;
conv2d_wgrad = "conv2d_wgrad(" , "kernel_h=" , INTEGER , "," , "kernel_w=" , INTEGER , ")" ;

conv1d_fprop = "conv1d_fprop(" , "kernel_w=" , INTEGER , ")" ;
depthwise_conv1d = "depthwise_conv1d(" , "kernel_w=" , INTEGER , ")" ;
group_conv1d = "group_conv1d(" , "kernel_w=" , INTEGER , "," , "groups=" , INTEGER , ")" ;

conv3d_fprop = "conv3d_fprop(" , "kernel_d=" , INTEGER , "," , "kernel_h=" , INTEGER , "," , "kernel_w=" , INTEGER , ")" ;
conv3d_dgrad = "conv3d_dgrad(" , "kernel_d=" , INTEGER , "," , "kernel_h=" , INTEGER , "," , "kernel_w=" , INTEGER , ")" ;
conv3d_wgrad = "conv3d_wgrad(" , "kernel_d=" , INTEGER , "," , "kernel_h=" , INTEGER , "," , "kernel_w=" , INTEGER , ")" ;

depthwise_conv2d = "depthwise_conv2d(" , "kernel_h=" , INTEGER , "," , "kernel_w=" , INTEGER , ")" ;
group_conv2d = "group_conv2d(" , "kernel_h=" , INTEGER , "," , "kernel_w=" , INTEGER , "," , "groups=" , INTEGER , ")" ;
group_conv3d = "group_conv3d(" , "kernel_d=" , INTEGER , "," , "kernel_h=" , INTEGER , "," , "kernel_w=" , INTEGER , "," , "groups=" , INTEGER , ")" ;

(* CONFIGURATION - all use unquoted identifiers *)
configuration = dtype_config | layout_config | arch_config | tile_config | stages_config 
              | alignment_config | cluster_config | swizzle_config | scheduler_config | scaling_config
              | iterator_config | split_k_config | operand_swap_config ;

dtype_config = ".with_dtype(" , "input=" , DTYPE , "," , "acc=" , DTYPE , "," , "output=" , DTYPE , ")" ;
layout_config = gemm_layout_config | conv_layout_config ;
gemm_layout_config = ".with_layout(" , "A=" , GEMM_LAYOUT , "," , "B=" , GEMM_LAYOUT , "," , "C=" , GEMM_LAYOUT , ")" ;
conv_layout_config = ".with_layout(" , "input=" , CONV_LAYOUT , "," , "filter=" , CONV_LAYOUT , "," , "output=" , CONV_LAYOUT , ")" ;

arch_config = ".with_arch(" , ARCH , ")" ;
tile_config = (".with_tile(" | ".with_threadblockshape(") , "m=" , INTEGER , "," , "n=" , INTEGER , "," , "k=" , INTEGER , ")" ;
stages_config = ".with_stages(" , INTEGER , ")" ;
alignment_config = ".with_alignment(" , "A=" , INTEGER , "," , "B=" , INTEGER , "," , "C=" , INTEGER , ")" ;
cluster_config = ".with_cluster(" , "m=" , INTEGER , "," , "n=" , INTEGER , "," , "k=" , INTEGER , ")" ;
swizzle_config = ".with_swizzle(" , "pattern=" , SWIZZLE , ")" ;
scheduler_config = ".with_scheduler(" , [ "tile=" , TILE_SCHEDULER ] , 
                   [ "," , "kernel=" , KERNEL_SCHEDULE ] , 
                   [ "," , "epilogue=" , EPILOGUE_SCHEDULE ] ,
                   [ "," , "splits=" , INTEGER ] , ")" ;
scaling_config = ".with_scaling(" , "alpha=" , FLOAT , "," , "beta=" , FLOAT , ")" ;

iterator_config = ".with_iterator(" , ITERATOR , ")" ;
split_k_config = ".with_split_k(" , "mode=" , SPLIT_K , "," , "slices=" , INTEGER , ")" ;
operand_swap_config = ".with_operand_swap(" , BOOL , ")" ;

(* EPILOGUE *)
epilogue = ">>" , epilogue_op ;
epilogue_op = simple_activation | parametric_activation | broadcast_op | fusion_op | custom_op ;

simple_activation = "relu()" | "gelu()" | "silu()" | "mish()" | "sigmoid()" | "tanh()" | "hardswish()" ;

parametric_activation = "leaky_relu(" , "alpha=" , FLOAT , ")" 
                      | "elu(" , [ "alpha=" , FLOAT ] , ")"
            | "clip(" , "min=" , FLOAT , "," , "max=" , FLOAT , ")"
                      | "clamp(" , "min=" , FLOAT , "," , "max=" , FLOAT , ")" ;

broadcast_op = "bias()" | "per_channel_scale()" | "per_row_scale()" | "per_col_scale()"
             | "row_broadcast(" , IDENTIFIER , ")" | "column_broadcast(" , IDENTIFIER , ")" ;

fusion_op = "scale(" , "alpha=" , FLOAT , ")"
          | "aux_store(" , IDENTIFIER , ")" | "aux_load(" , IDENTIFIER , ")" ;

(* Custom requires quotes for free-form expressions *)
custom_op = "custom(" , STRING , [ "," , "inputs=" , input_dict ] , ")" ;
input_dict = "{" , input_pair , { "," , input_pair } , "}" ;
input_pair = STRING , ":" , STRING ;

(* TERMINALS - unquoted identifiers *)
DTYPE = "fp64" | "float64" | "fp32" | "float32" | "fp16" | "float16" 
      | "bf16" | "bfloat16" | "tf32"
      | "fp8_e4m3" | "e4m3" | "fp8_e5m2" | "e5m2" 
      | "int8" | "s8" | "int16" | "s16" | "int32" | "s32"
      | "uint8" | "u8" | "uint16" | "u16" | "uint32" | "u32" ;

ARCH = "sm_100" | "sm100" | "sm_90a" | "sm90a" | "sm_90" | "sm90" 
     | "sm_89" | "sm89" | "sm_86" | "sm86" | "sm_80" | "sm80" | "sm_70" | "sm70" ;

GEMM_LAYOUT = "RowMajor" | "ColumnMajor" ;
CONV_LAYOUT = "TensorNHWC" | "TensorNDHWC" ;
SWIZZLE = "Identity1" | "Identity2" | "Identity4" | "Identity8" | "StreamK" ;
TILE_SCHEDULER = "default" | "persistent" | "stream_k" | "streamk" ;
KERNEL_SCHEDULE = "auto" | "cp_async" | "cp_async_cooperative" | "tma" | "tma_cooperative" | "tma_pingpong" ;
EPILOGUE_SCHEDULE = "auto" | "tma" | "tma_cooperative" | "no_smem" ;
ITERATOR = "analytic" | "optimized" | "fixed_channels" | "few_channels" | "fixed_stride_dilation" ;
SPLIT_K = "none" | "serial" | "parallel" ;
BOOL = "true" | "false" ;

INTEGER = DIGIT , { DIGIT } ;
FLOAT = [ "-" ] , INTEGER , [ "." , INTEGER ] ;
STRING = "'" , { ANY_CHAR - "'" } , "'" ;
IDENTIFIER = LETTER , { LETTER | DIGIT | "_" } ;
DIGIT = "0" | "1" | "2" | "3" | "4" | "5" | "6" | "7" | "8" | "9" ;
LETTER = "a" .. "z" | "A" .. "Z" ;

(* CONSTRAINTS (compiler-enforced):
 *
 * REQUIRED: .with_dtype(), .with_arch(), .with_layout() (GEMM only)
 *
 * ARCH-GATED:
 *   SM70-89: .with_tile(), .with_swizzle(), .with_iterator()
 *   SM90+:   .with_threadblockshape(), .with_cluster(), .with_scheduler(), .with_operand_swap()
 *   SM80+: bf16 | SM90+: fp8_e4m3, fp8_e5m2
 *
 * SM90+ CONSTRAINTS (compiler-enforced):
 *   1. ALWAYS use sm_90a (not sm_90) - the "a" enables wgmma/warp-specialized features
 *      - Applies to ALL schedules: tma, tma_cooperative, cp_async, etc.
 *   2. Use .with_threadblockshape() NOT .with_tile() - compiler rejects .with_tile() on SM90+
 *   3. TMA alignment: (alignment * element_size) % 16 == 0
 *      - fp16/bf16: A/B/C >= 8 (8 * 2 = 16 bytes)
 *      - fp32: A/B/C >= 4 (4 * 4 = 16 bytes)
 *   4. kernel=tma_cooperative requires epilogue=tma_cooperative (or auto)
 *      - Mismatch causes "MMA_TILE_M must divide EPI_TILE_M" error
 *   5. Cooperative kernels: tile_m / cluster_m >= 128
 *      - Example FAIL: m=128 + cluster_m=2 → per_cta_m=64 < 128
 *   6. TMA cooperative REQUIRES explicit .with_stages(2) - compiler rejects missing stages
 *      - stages = (228KB - epilogue_smem - 8KB) / per_stage_smem
 *      - Large fp32 tiles exhaust smem (e.g., 256×128×64 fp32 → only 1 stage)
 *      - Fix: add .with_stages(2), use smaller tile, fp16/bf16, or kernel='tma'
 *   7. .with_operand_swap(true) REQUIRES square output (M == N) - runtime check enforced
 *      - Uses (A@B)^T = B^T @ A^T identity for zero-overhead layout reinterpretation
 *      - Enables RS GMMA variant (lower SMEM) and larger clusters for FP32 GEMM
 *      - Column-major [N,M] == Row-major [M,N] only when M == N
 *      - Note: FP16/BF16 already use RS GMMA with RowMajor B; operand_swap is FP32-specific benefit
 *
 * SM90+ GEMM template:
 *   gemm().with_dtype(input=fp16, acc=fp32, output=fp16)
 *     .with_layout(A=RowMajor, B=ColumnMajor, C=RowMajor).with_arch(sm_90a)
 *     .with_threadblockshape(m=256, n=128, k=64).with_alignment(A=8, B=8, C=8)
 *     .with_scheduler(kernel=tma_cooperative, epilogue=tma_cooperative)
 *
 * SM90+ FP32 GEMM with operand swap (square matrices only):
 *   gemm().with_dtype(input=fp32, acc=fp32, output=fp32)
 *     .with_layout(A=RowMajor, B=RowMajor, C=RowMajor).with_arch(sm_90a)
 *     .with_threadblockshape(m=128, n=128, k=32).with_cluster(m=4, n=2, k=1)
 *     .with_alignment(A=4, B=4, C=4).with_stages(3).with_operand_swap(true)
 *)
\end{lstlisting}

\subsection{Example SOL Report}
\label{sec:appendix:sol-report}

We include a sample SOL report for KernelBench Problem~001 (4096$\times$4096 FP32 GEMM) here.
The main body uses FP32/TF32 assumptions matching the problem's declared precision.
The FP16 augmentation at the end shows how the bounds tighten when reduced-precision Tensor Cores are assumed, and is used for scheduling and integrity checking.

\begin{lstlisting}[style=ebnf,caption={SOL report for Problem~001 (4096$\times$4096 FP32 GEMM).},label={lst:sol-report-001},basicstyle=\ttfamily\tiny]
# Speed-of-Light (SOL) Analysis

## 1. Problem Characterization
We analyze the reference op: square GEMM
C = A x B, where A,B,C in R^{NxN}, with N = 4096 and FP32 inputs.

### FLOP count (lower bound)
For dense GEMM, each output element is a dot product of length N:
- per C_{ij}: N multiplies + (N-1) adds ~ 2N FLOPs (2 FLOPs per MAC).
- total elements: N^2

Total FLOPs = 2N^3

Plug in N=4096:
- 4096^3 = 68,719,476,736
- 2 x 4096^3 = 137,438,953,472

=> Total FLOPs = 1.374e+11 FLOPs

### Global memory bytes (best-case lower bound)
Best-case DRAM traffic assumes:
- A is read once (each element brought from DRAM at least once)
- B is read once
- C is written once
- No extra reads/writes for intermediates (none exist beyond GEMM)

Each matrix has N^2 elements, FP32 = 4 bytes/element.

N^2 = 4096^2 = 16,777,216 elements
Bytes per matrix: 16,777,216 x 4 = 67,108,864 bytes
Total bytes (2 reads + 1 write): 3 x 67,108,864 = 201,326,592 bytes

=> Total bytes = 2.013e+08 bytes (~192 MiB)

### Arithmetic intensity
AI = FLOPs / Bytes = 1.374e+11 / 2.013e+08 ~ 682.6 FLOPs/byte

### TF32 note for FP32 GEMM
Inputs are FP32 and this is GEMM. Given the runtime context:
- torch.backends.cuda.matmul.allow_tf32: True
- torch.get_float32_matmul_precision(): highest

We assume PyTorch uses TF32 Tensor Cores on H100 (SM90) for matmul.
SOL therefore uses TF32 Tensor Core peak throughput.

---

## 2. Hardware Limits (Clock-aware)
GPU: NVIDIA H100 80GB HBM3 (Hopper, SM90)

Locked application clock:
- SM clock: 1500 MHz (locked application clock for benchmarking)
- Max SM clock: 1980 MHz
- Memory clock: 2619 MHz (max)

Peak specs (H100 SXM 80GB, at max clocks, dense, no sparsity):
- TF32 Tensor Core peak: 494.7 TFLOP/s (dense; the 989 TFLOP/s figure
  is with structured sparsity)
- FP16 Tensor Core peak: 989.4 TFLOP/s (dense; 1979 TFLOP/s is sparse)
- HBM3 peak bandwidth: 3.35 TB/s = 3350 GB/s

Clock-scaling (linear with clock ratio):

SM clock ratio:      1500 / 1980 = 0.7576
Effective TF32 peak: 494.7 TFLOP/s x 0.7576 = 374.77 TFLOP/s
Effective FP16 peak: 989.4 TFLOP/s x 0.7576 = 749.55 TFLOP/s

Memory clock ratio:  2619 / 2619 = 1.0
Effective bandwidth: 3.35 TB/s x 1.0 = 3.35 TB/s

=> Effective peak TF32:     ~374.77 TFLOP/s
=> Effective peak FP16:     ~749.55 TFLOP/s
=> Effective peak bandwidth: ~3.35 TB/s

---

## 3. Theoretical Minimum Time (TF32)

Compute-bound time:
  T_compute = FLOPs / Peak FLOP/s
            = 1.374e+11 / 374.77e+12
            = 0.367 ms

Memory-bound time:
  T_mem = Bytes / Peak BW
        = 2.013e+08 / 3.35e+12
        = 0.060 ms

SOL = max(T_compute, T_mem) = max(0.367, 0.060) = 0.367 ms
Primary bottleneck: Compute-bound.

---

## 4. Roofline Analysis

Ridge point = Peak FLOP/s / Peak BW
            = 374.77e+12 / 3.35e+12
            = 111.9 FLOPs/byte

Kernel AI ~ 682.6 FLOPs/byte >> Ridge ~ 111.9 FLOPs/byte
=> Compute-bound region on the roofline plot.

---

## 5. Summary
- Operation: FP32 GEMM (assumed TF32 Tensor Cores)
- Total work: 2N^3 = 1.374e+11 FLOPs for N=4096
- Best-case DRAM traffic: 2.013e+08 bytes (~192 MiB)
- Arithmetic intensity: ~682.6 FLOPs/byte
- Clock-aware effective peaks (at 1500 MHz locked clock):
  - TF32 compute: ~374.77 TFLOP/s (dense, scaled by 1500/1980)
  - HBM bandwidth: ~3.35 TB/s
- Lower-bound times:
  - Compute time: ~0.367 ms
  - Memory time:  ~0.060 ms

=> Theoretical minimum execution time (SOL): ~0.367 ms
=> Primary bottleneck: Compute throughput

===================================================================
# FP16 Augmentation
===================================================================

Precision Assumption:
  Kernel casts FP32 data to FP16 on-chip and uses FP16 Tensor Cores
  (2x throughput). Inputs, outputs, and weights remain FP32 in DRAM
  - memory traffic is unchanged.

- Compute peak: FP16 TC dense (749.55 TFLOP/s at 1500 MHz,
                scaled from 989.4 TFLOP/s at 1980 MHz)
- Memory traffic: FP32 data (4 bytes/element) - NOT halved,
                  because inputs/outputs enter and leave DRAM as FP32
- HBM bandwidth: 3350.0 GB/s

Work Estimates:
- Total FLOPs:        1.3744e+11 (unchanged)
- Total bytes (FP32): 2.0133e+08 (unchanged from TF32 SOL)

Theoretical Minimum Times:
              TF32 (dense)  FP16 (dense)
  Peak        374.77        749.55        TFLOP/s
  Compute     0.3667 ms     0.1834 ms
  Memory      0.0601 ms     0.0601 ms
  SOL         0.3670 ms     0.1834 ms
  Bottleneck  compute       compute

FP16/TF32 ratio: 0.500x (pure 2x compute speedup; memory-bound
unchanged because DRAM I/O remains FP32)

===================================================================
# Structured JSON Output
===================================================================

{
  "problem_id": "001",
  "timestamp": "2026-01-27 19:22:15",
  "total_flops": 137438953472,
  "total_bytes": 201326592,
  "theoretical_runtime_ms": 0.367,
  "sm_clock_mhz": 1500.0,
  "peak_type": "TF32 TC (dense)",
  "peak_tflops_effective": 374.77,
  "theoretical_runtime_ms_fp16": 0.1834,
  "fp16_peak_tflops_effective": 749.55,
  "fp16_note": "FP16 compute (2x TC throughput), FP32 memory (inputs/outputs/weights stay FP32 at DRAM boundary)",
  "details": {
    "full_analysis": "<see TF32 markdown report above>"
  }
}
\end{lstlisting}

\subsection{KernelBench LLM-Relevant Problem Subset}
\label{sec:appendix:kernelbench-subset}


\paragraph{Included KernelBench IDs (Levels 1--3).}
\noindent\textbf{L1:} 1, 2, 3, 4, 6, 7, 8, 9, 16, 17, 18, 21, 22, 23, 25, 26, 36, 40, 47, 48, 67, 76, 86, 87, 88, 89, 90, 91, 92, 95, 97. \\
\textbf{L2:} 9, 28, 29, 37, 40, 41, 53, 56, 59, 62, 63, 66, 70, 76, 81, 86, 88, 94, 97, 99. \\
\textbf{L3:} 1, 2, 3, 43, 44, 48, 49, 50.

\begin{table*}[t]
  \centering
  \scriptsize
  \setlength{\tabcolsep}{3pt}
  \renewcommand{\arraystretch}{0.92}
  \caption{KernelBench problems in our LLM-relevant subset (Levels 1--3).}
  \label{tab:kernelbench-subset}
  \begin{tabular}{@{}c c p{0.78\textwidth}@{}}
    \toprule
    Level & ID & Rationale for inclusion \\
    \midrule
    L1 & 1  & Basic GEMM baseline. \\
    L1 & 2  & LLM-like GEMM shapes (e.g., M=2048, K=8192, N=4096). \\
    L1 & 3  & Batched matmul (BMM) used in attention score/value products. \\
    L1 & 4  & Matrix--vector multiply representative of single-token decode. \\
    L1 & 6  & Matmul with large \(K\) (common in MLP projections). \\
    L1 & 7  & Matmul with small \(K\) (e.g., attention head dimension). \\
    L1 & 8  & Irregular shapes (non power-of-2) that occur in practice. \\
    L1 & 9  & Tall-skinny matmul (prefill with long sequences). \\
    L1 & 16 & Transposed-\(A\) layout variant common in GEMM calls. \\
    L1 & 17 & Transposed-\(B\) layout variant common for weight matrices. \\
    L1 & 18 & Both operands transposed (layout coverage). \\
    L1 & 21 & Sigmoid for gating patterns (e.g., GLU-style gates). \\
    L1 & 22 & Tanh used in some gating/activation variants. \\
    L1 & 23 & Softmax (core attention primitive). \\
    L1 & 25 & SiLU/Swish (dominant MLP activation in many modern LLMs). \\
    L1 & 26 & GELU (used in GPT-2/BERT and some contemporary models). \\
    L1 & 36 & RMSNorm (dominant normalization in modern decoder LLMs). \\
    L1 & 40 & LayerNorm (still used in many transformer variants). \\
    L1 & 47 & Sum reduction used inside normalization and statistics. \\
    L1 & 48 & Mean reduction used inside LayerNorm and statistics. \\
    L1 & 67 & 1D convolution used in SSM/long-conv text models. \\
    L1 & 76 & Dilated/strided 1D conv variant for hierarchical SSM designs. \\
    L1 & 86 & Depthwise-separable conv (efficient channel-wise processing). \\
    L1 & 87 & Pointwise conv (channel mixing / fusion proxy). \\
    L1 & 88 & Fast GELU approximation (common fused activation variant). \\
    L1 & 89 & Cumsum (prefix-scan) used in SSM/linear-attention recurrences. \\
    L1 & 90 & Cumprod used in some state-space dynamics. \\
    L1 & 91 & Exclusive cumsum variant (scan coverage). \\
    L1 & 92 & Reverse cumsum variant (reverse-time scan coverage). \\
    L1 & 95 & Cross-entropy loss (standard LLM training objective). \\
    L1 & 97 & Scaled dot-product attention (maps to FlashAttention in practice). \\
    L2 & 9  & Fused matmul + elementwise (proxy for epilogue and MLP fusions). \\
    L2 & 28 & BMM fusion representative of multi-head attention dataflow. \\
    L2 & 29 & Fused linear + activation (MLP fusion pattern). \\
    L2 & 37 & Fused linear + normalization (proxy for norm-adjacent fusions). \\
    L2 & 40 & Fused linear + residual add (transformer block core pattern). \\
    L2 & 41 & GEMM + multi-activation fusion (MLP epilogue diversity). \\
    L2 & 53 & GEMM + activation fusion (covers activation/scaling variants). \\
    L2 & 56 & Matmul + gating + reduction (proxy for gated aggregation patterns). \\
    L2 & 59 & Matmul + SiLU/Swish + scaling (common MLP fusion). \\
    L2 & 62 & Matmul + normalization + activation (fused post-linear processing). \\
    L2 & 63 & GEMM + ReLU + divide (activation + scaling fusion). \\
    L2 & 66 & Attention-like fusion with dropout (training attention pattern). \\
    L2 & 70 & GEMM + sigmoid gate + residual add (SwiGLU-like gating proxy). \\
    L2 & 76 & GEMM + bias add + ReLU (classic epilogue fusion). \\
    L2 & 81 & Complex epilogue fusion with Swish (stress fused elementwise). \\
    L2 & 86 & Matmul + divide + GELU (MLP fusion with scaling). \\
    L2 & 88 & SwiGLU-like gated fusion (common LLM MLP pattern proxy). \\
    L2 & 94 & Expert MLP proxy: GEMM + bias/activation + normalization. \\
    L2 & 97 & Matmul + bias + norm + Swish (fused post-linear processing). \\
    L2 & 99 & Attention-like fusion (matmul + GELU + softmax). \\
    L3 & 1  & MLP (basic feedforward block). \\
    L3 & 2  & Shallow wide MLP (LLM FFN-like width). \\
    L3 & 3  & Deep narrow MLP (depth/width trade-off). \\
    L3 & 43 & Causal attention block (core decoder attention). \\
    L3 & 44 & Full GPT block (attention + FFN). \\
    L3 & 48 & Mamba SSM block (emerging text SSM architecture). \\
    L3 & 49 & Mamba SSM with state output (streaming/stateful variant). \\
    L3 & 50 & ReLU self-attention variant (alternative attention formulation). \\
    \bottomrule
  \end{tabular}
\end{table*}

\end{document}